\documentclass[runningheads]{llncs}

\usepackage{eccv}

\usepackage{eccvabbrv}

\usepackage{graphicx}
\usepackage{booktabs}

\usepackage[accsupp]{axessibility}  
\usepackage{times}
\usepackage{soul}
\usepackage{url}
\usepackage[utf8]{inputenc}
\usepackage{amsmath}
\usepackage{booktabs}
\usepackage{algorithm}
\usepackage{algorithmic}

\usepackage{amsmath,amsfonts,bm}

\def\eqref#1{equation~\ref{#1}}

\def\1{\bm{1}}

\DeclareMathAlphabet{\mathsfit}{\encodingdefault}{\sfdefault}{m}{sl}
\SetMathAlphabet{\mathsfit}{bold}{\encodingdefault}{\sfdefault}{bx}{n}

\usepackage{multirow}
\usepackage{amssymb}
\usepackage{subcaption} 
\usepackage{pifont}

\usepackage{xcolor}

\usepackage{hyperref}

\usepackage{orcidlink}

\begin{document}

\title{Diversity-Aware Agnostic Ensemble of\\ Sharpness Minimizers} 

\titlerunning{Diversity-Aware Agnostic Ensemble of Sharpness Minimizers}

\author{Anh Bui\inst{1}$^*$ \and
Vy Vo\inst{1}$^*$ \and
Tung Pham\inst{2} \and
Dinh Phung\inst{1} \and
Trung Le\inst{1}}

\authorrunning{A. Bui et al.}

\institute{Monash University, Australia \and
VinAI Research, Vietnam \\
\email{\{tuananh.bui,tran.vo,trunglm, dinh.phung\}@monash.edu, v.tungph4@vinai.io}}

\maketitle

\def\thefootnote{*}\footnotetext{Equal contribution}\def\thefootnote{\arabic{footnote}}

\begin{abstract}
There has long been plenty of theoretical and empirical evidence supporting the success of ensemble learning. Deep ensembles in particular take advantage of training randomness and expressivity of individual neural networks to gain prediction diversity, ultimately leading to better generalization, robustness and uncertainty estimation. In respect of generalization, it is found that pursuing wider local minima result in models being more robust to shifts between training and testing sets. A natural research question arises out of these two approaches as to whether a boost in generalization ability can be achieved if ensemble learning and loss sharpness minimization are integrated. Our work investigates this connection and proposes DASH - a learning algorithm that promotes diversity and flatness within deep ensembles. More concretely, DASH encourages base learners to move divergently towards low-loss regions of minimal sharpness. We provide a theoretical backbone for our method along with extensive empirical evidence demonstrating an improvement in ensemble generalizability. 
\end{abstract}

\section{Introduction}
Ensemble learning refers to learning a combination of multiple models in a way that the joint performance is better than than any of the ensemble members (so-called base learners). 
An ensemble can be an explicit collection of functionally independent models where the final decision is formed via approaches like averaging or majority voting of individual predictions. 
It can implicitly be a single model subject to stochastic perturbation of model architecture during training ~\cite{srivastava2014dropout,wan2013regularization} or composed of sub-modules sharing some of the model parameters ~\cite{wenzel2020hyperparameter,wen2020batchensemble}. An ensemble is called \textit{homogeneous} if its base learners belong to the same model family or architecture and \textit{heterogeneous} otherwise.

Traditional bagging technique ~\cite{breiman1996bagging} is shown to reduce variance among the base learners while boosting methods ~\cite{breiman1996bias,zhang2008rotboost} are more likely to help reduce bias and improve generalization. Empirical evidence further points out that ensembles perform at least equally well as their base learners ~\cite{krogh1994neural} and are much less fallible when the members are independently erroneous in different regions of the feature space ~\cite{hansen1990neural}. Deep learning models in particular often land at different local minima valleys due to with training randomness, from initializations, mini-batch sampling, etc. This causes disagreement on predictions among model initializations given the same input. Meanwhile, deep ensembles (i.e., ensembles of deep neural networks) are found to be able to ``smooth out'' the highly non-convex loss surface, resulting in a better predictive performance ~\cite{hansen1990neural,perrone1995networks,garipov2018loss,fort2019emergent,li2018visualizing}. Ensemble models also benefit from the enhanced diversity in predictions, which is highlighted as another key driving force behind the success of ensemble learning ~\cite{dietterich2000ensemble}. Further studies suggest that higher diversity among base learners leads to better robustness and predictive performance ~\cite{hansen1990neural,ovadia2019can,fort2019deep,sinha2020diversity}. A recent work additionally shows that deep ensembles in general yield the best calibration under dataset shifts ~\cite{ovadia2019can}. 
  
Tackling model generalization from a different approach, sharpness-aware minimization is a line of work that seeks the minima within the flat loss regions,  along which SAM ~\cite{foret2021sharpnessaware} is the most popular method.  Flat minimizers have been theoretically and empirically proven in various applications to yield better testing accuracies ~\cite{JiangNMKB20,PetzkaKASB21,DziugaiteR17}. 
At every training step, SAM performs one gradient ascent step to find the worst-case perturbations on the parameters. 
Given plenty of advantages of ensemble models, a natural question thus arises as to whether ensemble learning and sharpness-aware minimization can be integrated to boost model generalizability. 
In other words, \textit{can we learn a deep ensemble of sharpness minimizers such that the entire ensemble is more generalizable?}

Motivated by this connection, our work proposes to improve generalization performance by learning an ensemble of deep sharpness-aware learners. We first develop a theory showing that the general loss of the ensemble can be reduced by minimizing loss sharpness in both the ensemble and its base learners (See \cref{thm:simple_theorem}). Our theoretical development sheds lights on how to guide individual learners in a deep ensemble to be well-behaved and collaborate effectively on a high-dimensional loss landscape. 

In addition to generalization, we also target other desiderata of ensemble learning including diversity, robustness and low uncertainty. While the endeavor to address all desiderata within a single framework might be ambitiously challenging, past studies suggest that fostering diversity among the base learners is a multi-purpose approach that can lead to improved generalizability ~\cite{li2012diversity,sun2018structural,ortega2022diversity}, stability ~\cite{zhang2020diversified}, and adversarial robustness ~\cite{pang2019improving} in the ensemble. To this end, we contribute a novel agnostic diversity-aware constraint that aims to navigate the individual learners to explore multiple wide minima in a divergent fashion. The diversity-aware term attempts to minimize the pairwise $\textrm{KL}$ divergence among the base learners. Such a term is \textit{agnostic} in the sense that it is introduced early on in the process of searching for the perturbed model. Intuitively, we expect the term to "look ahead" for potential gradient pathways that would guide the updated model to satisfy the goal. 

In summary, our contributions in this paper are summarized as follows: \\
\ding{192} We propose \textbf{DASH} Ensemble - an ensemble learning method for \textbf{D}iversity-aware \textbf{A}gnostic Ensemble of \textbf{Sh}arpness Minimizers. \textbf{DASH} seeks to minimize generalization loss by  directing the ensemble and its base classifiers towards diverse loss regions of maximal flatness. \\
\ding{193} We provide a theoretical development for our method, followed by the technical insights into how adding the agnostic diversity-aware term helps introduce diversity in the ensemble and results in better predictive performance and uncertainty estimation capability than the baseline methods. \\
\ding{194} Across various image classification tasks, we demonstrate an improvement in model generalization capacity of both homogeneous and heterogeneous ensembles up to $6\%$, where the latter benefits significantly.

\section{Related works}
\paragraph{Ensemble Learning.} The rise of ensemble learning dates back to the development of classical techniques like bagging ~\cite{breiman1996bagging} or boosting ~\cite{breiman1996bias,freund1996experiments,friedman2001greedy,zhang2008rotboost} for improving model generalization. 
While bagging algorithm involves training independent weak learners in parallel, boosting methods iteratively combine base learners to create a strong model where successor learners try to correct the errors of predecessor ones. 
In the era of deep learning, there has been an increase in attention towards ensembles of deep neural networks. 
A deep ensemble made up of low-loss neural learners has been consistently shown to yield to outperform an individual network ~\cite{hansen1990neural,perrone1995networks,huang2017snapshot,garipov2018loss,evci2020rigging}. 
In addition to predictive accuracy, deep ensembles has achieved successes in such other areas as uncertainty estimation ~\cite{lakshminarayanan2017simple,ovadia2019can,gustafsson2020evaluating} 
or adversarial robustness ~\cite{pang2019improving,yang2021trs,yang2020dverge}. 

Ensembles often come with high training and testing costs that can grow linearly with the size of ensembles. 
This motivates recent works on efficient ensembles for reducing computational overhead without compromising their performance. 
One direction is to leverage the success of Dynamic Sparse Training ~\cite{liu2021sparse,evci2022gradient} 
to generate an ensemble of sparse networks with lower training costs while maintaining comparable performance with dense ensembles ~\cite{liudeep}. 
Another light-weight ensemble learning method is via pseudo or implicit ensembles that involves training a single model that exhibits the behavior or characteristic of an ensemble. 
Regularization techniques such as Drop-out ~\cite{srivastava2014dropout,gal2016dropout}, Drop-connect ~\cite{wan2013regularization} or Stochastic Depth ~\cite{huang2016deep} can be viewed as an ensemble network by masking the some units, connections or layers of the network.  
Other implicit strategies include training base learners with different hyperparameter configurations ~\cite{wenzel2020hyperparameter}, 
decomposing the weight matrices into individual weight modules for each base learners ~\cite{wen2020batchensemble} or using multi-input/output configuration to learn independent sub-networks within a single model ~\cite{havasi2020training}. 

\paragraph{Sharpness-Aware Minimization.} There has been a growing body of theoretical and empirical studies on the connection between loss sharpness and generalization capacity ~\cite{DBLP:conf/nips/HochreiterS94, neyshabur2017exploring, dinh2017sharp, fort2019emergent}. Convergence in flat regions of wider local minima has been found to improve out-of-distribution robustness of neural networks ~\cite{JiangNMKB20, PetzkaKASB21, DziugaiteR17}. Some other works ~\cite{DBLP:conf/iclr/KeskarMNST17,Jastrzebski2017ThreeFI, wei2020implicit} study the effect of the covariance of gradient or training configurations such as batch size, learning rate, dropout rate on the flatness of minima. One way to encourage search in flat minima is by adding regularization terms to the loss function such as Softmax output's low  entropy penalty ~\cite{DBLP:conf/iclr/PereyraTCKH17,Chaudhari2017EntropySGDBG} or distillation losses ~\cite{Zhang2018DeepML,zhang2019your}. 

SAM ~\cite{foret2021sharpnessaware} is a recent flat minimizer widely known for its effectiveness and scalability, which encourages the model to search for parameters in the local regions that are uniformly low-loss. SAM has been actively exploited in various applications: meta-learning bi-level optimization in ~\cite{abbas2022sharp}, federated learning ~\cite{qu2022generalized}, domain generalization ~\cite{cha2021swad}, multi-task learning ~\cite{phan2022improving} or for vision transformers ~\cite{chen2021vision} and language models ~\cite{bahri-etal-2022-sharpness}. Coming from two different directions, ensemble learning and sharpness-aware minimization yet share the same goal of improving generalization. Leveraging these two powerful learning strategies in a single framework remains underexplored. Our work contributes an effort to fill in this research gap.

\section{Proposed method}
In this section, we first present the theoretical development demonstrating why sharpness-aware ensemble learning is beneficial for improving the generalization of ensemble models. We later introduce how to promote ensemble diversity by enforcing a novel agnostic diversity-aware constraint among the base learners.

\paragraph{\textbf{Ensemble Setting and Notations.}} We first describe the ensemble setting and the notations used throughout our paper. Given $m$ base learners $f_{\theta_{i}}^{(i)}\left(x\right),i=1,...,m$,
we define the ensemble model
\[
f_{\theta}^{\mathrm{ens}}\left(x\right)=\frac{1}{m}\sum_{i=1}^{m}f_{\theta_{i}}^{(i)}\left(x\right),
\]
where $\theta_{\mathrm{ens}}=\left[\theta_{i}\right]_{i=1}^m$, 
$x\in\mathbb{R}^{d}$, and $f\left(x\right)\in\Delta_{M-1}=\left\{ \pi\in\mathbb{R}^{M}:\pi\geq0\land\Vert \pi\Vert _{1}=1\right\} $. Here $\theta_i$ and $\theta_{\mathrm{ens}}$ denote the parameters w.r.t the classifier $f_{\theta_i}^{(i)}$ and the ensemble classifier $f_{\theta}^{\mathrm{ens}}$, respectively. Note that the base learners $f_{\theta_{i}}^{(i)}$ can have different architectures.

Assume that $\ell:\mathbb{R}^M\times\mathcal{Y}\xrightarrow{}\mathbb{R}$, where
$\mathcal{Y}=\left[M\right]= \{1,\dots,M\}$ is the label set, is a convex and
bounded loss function. The training set  is denoted by $\mathcal{S}=\left\{ \left(x_{i},y_{i}\right)\right\} _{i=1}^{N}$
of data points $\left(x_{i},y_{i}\right)\sim\mathcal{D}$, where $\mathcal{D}$ is
a data-label distribution. We denote 
$\mathcal{L}_{\mathcal{S}}\left(\theta_{i}\right) = \frac{1}{N}\sum_{j=1}^{N}\ell\left(f_{\theta_{i}}^{i}\left(x_{j}\right),y_{j}\right)$ and 
$\mathcal{L}_{\mathcal{D}}\left(\theta_{i}\right) = \mathbb{E}_{(x,y)\sim \mathcal{D}}\big[ \ell(f_{\theta_i}^i(x),y)\big]$ 
as the empirical and general losses w.r.t. the base learner $\theta_i$, respectively.

Similarly, for the ensemble model, we respectively define the empirical and general losses as 
$\mathcal{L}_{\mathcal{S}}\left(\theta_{\text{ens}}\right) = \frac{1}{N}\sum_{j=1}^{N}\ell\left(f_{\theta}^{\mathrm{ens}}\left(x_{j}\right),y_{j}\right)$ and
$\mathcal{L}_{\mathcal{D}}\left(\theta_{\text{ens}}\right) = \mathbb{E}_{(x,y)\sim \mathcal{D}}\big[ \ell(f_{\theta}^{\mathrm{ens}}(x),y)\big]$.

\subsection{Sharpness-aware Ensemble Learning} \label{sec:sa_ens}

\paragraph*{Standard Sharpness-Aware Minimization.} As introduced in SAM \cite{foret2021sharpnessaware}, given a single model $f_\theta$, the generalization error 
$\mathcal{L}_{\mathcal{D}}(\theta)$ can be upper-bounded by the sharpness of the model, i.e., $\max_{\theta':\Vert\theta'-\theta\Vert\leq \rho}\mathcal{L}_{\mathcal{S}}\left(\theta'\right) - \mathcal{L}_{\mathcal{S}}\left( \theta \right)$, where $\rho > 0$ is the perturbed radius. 
More specifically, Theorem 1 in ~\cite{foret2021sharpnessaware} shows that

\begin{equation}
\mathcal{L}_{\mathcal{D}}(\theta) \leq  \max_{\theta':\Vert\theta'-\theta\Vert\leq \rho}\mathcal{L}_{\mathcal{S}}\left(\theta'\right) + h ( \| \theta \|_2^2 / \rho), \nonumber
\end{equation}

where $h$ is a strictly increasing function that depends on $\theta$ and $\rho$. 
The theorem suggests that minimizing the sharpness of a single model can improve its generalizability. Upon the success of SAM, many consecutive works have been proposed to improve the sharpness-aware minimization. However, they are all limited to a single model.

\paragraph*{Sharpness-Aware Ensemble learning.} We now present a sharpness-aware upper bound for the general loss of the ensemble model. To assist readability, we provide the \textbf{simplified} version in the following theorem. The full development can be found in the supplementary materials.

\begin{theorem}
    \label{thm:simple_theorem} Assume that the loss function $\ell$ is convex
    and upper-bounded by $L$.  
    With the probability at least $1-\delta$
    over the choices of $\mathcal{S}\sim\mathcal{D}^{N}$, for any $0\leq\gamma\leq1$,
    we have
    \begin{align}\label{eq:sharpness_ens}
    \mathcal{L}_{\mathcal{\mathcal{D}}}\left(\theta_{\mathrm{ens}}\right)&\leq \frac{\left(1-\gamma\right)}{m}\sum_{i=1}^{m}\max_{\theta_{i}^{'}:\Vert\theta_{i}^{'}-\theta_{i}\Vert \leq \rho}\mathcal{L}_{\mathcal{S}}\left(\theta_{i}^{'}\right) \\ \nonumber
    &+\gamma\max_{\theta^{\prime}_{\mathrm{ens}}:\Vert\theta^{\prime}_{\mathrm{ens}}-\theta_{\mathrm{ens}}\Vert \leq \sqrt{m}\rho} \mathcal{L}_{\mathcal{S}}(\theta^{\prime}_{\mathrm{ens}}) + \mathcal{H}(m, \{ \|\theta_i\|^2_2 / \rho \}_{i=1}^m ),
    \end{align}
    where $\mathcal{H}$ is a strictly increasing function of $m$, $\rho$ and set of model parameter $\{\theta_i\}_{i=1}^m$.
\end{theorem}

In the RHS of the inequality, the first term refers to the average sharpness of each independent base learner, while the second term focuses on the sharpness of the entire ensemble model. The trade-off coefficient $\gamma$ signifies the levels of sharpness-aware enforcement for the ensemble model alone and its base learners themselves. Our \cref{thm:simple_theorem} indicates that the generalization performance of the ensemble model can be improved by promoting the sharpness in both the entire ensemble as well as in the individual base learners.

\paragraph*{The dynamics of two modes of sharpness.} Intuitively, \cref{eq:sharpness_ens} suggests an effective ensemble dynamics where the base learners are not only encouraged to achieve good performance individual but also to contribute \textit{synergistically} to the ensemble. It is worth noting that the former may foster the latter behavior while the latter alone is likely to be insufficient. \cref{sec:di_ens} will later discuss one possible \textit{antagonistic} behavior within an ensemble. We now investigate how $\gamma$ should be optimally chosen by studying the impact of these two modes of sharpness on the ensemble performance. 

\begin{figure}
    \centering
    \includegraphics[width=0.6\columnwidth]{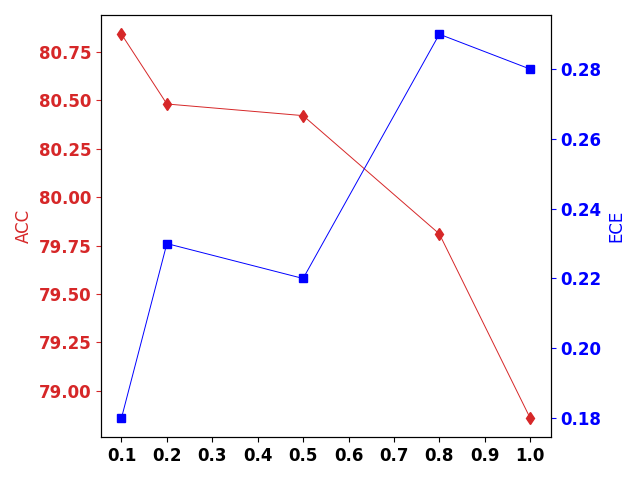}
    \caption{Tuning for hyper-parameter $\gamma$. Both the ensemble accuracy (ACC, higher is better) and the expected calibration error (ECE, lower is better) peak when $\gamma = 0.1$. See \cref{tab:tune-gamma} for other metrics.} 
    \label{fig:tuning-gamma}
\end{figure}

We conduct the experiments on the CIFAR100 dataset by varying $\gamma$ and observing the ensemble performance as shown in \cref{fig:tuning-gamma}. It can be seen that varying $\gamma$ does significantly affect the ensemble performance, with the difference of more than $1.8\%$ in ensemble accuracy. Interestingly, the ensemble accuracy and its uncertainty estimation capability peak at $\gamma = 0.1$ and decrease when $\gamma$ increases. This empirical observation confirms our intuition that to enhance the generalization ability of the ensemble model, one should focus more on minimizing the sharpness of the base learners than on minimizing the sharpness of the ensemble model. This observation concurs with the finding in ~\cite{allen2022towards} that the ensemble model's generalization ability is more sensitive to the sharpness of the base learners than the ensemble model itself.

\subsection{Diversity-Aware Agnostic 
Ensemble of Flat Base Learners}\label{sec:di_ens}
From the previous section, we have known that solely enforcing sharpness on the entire ensemble, that is to treat the ensemble as a single model and naively apply SAM, is not an optimal strategy. However, \cref{fig:tuning-gamma} also highlights that enforcing a larger degree of sharpness within individual learners does yield a positive collaborative effect. However, we argue that the current approach still has not maximized the synergy of the learners via this strategy alone. In the following, we provide a theoretical analysis for one potential antagonistic behavior of the base learners.   

For the current mini-batch $B$, we define 
\begin{align*}
    \widetilde{\mathcal{L}_{B}}(\theta_i) = \mathcal{L}_{B}(\theta_i) + \gamma \ \mathcal{L}^{\mathrm{ens}}_{B}(\theta_i, \theta_{\ne i}).
\end{align*}
When we enforce the sharpness within the learner $\theta_i$, the model $f^i_{\theta_i}$ is updated as
\begin{align}
\theta_{i}^{a} & =\theta_{i}+\rho_1\frac{\nabla_{\theta_{i}}\widetilde{\mathcal{L}_{B}}\left(\theta_{i}\right)}{\Vert\nabla_{\theta_{i}}\widetilde{\mathcal{L}_{B}}\left(\theta_{i}\right)\Vert} 
, \label{eq:SAM_update}\\
\theta_{i} & =\theta_{i}-\eta\nabla_{\theta_{i}}\widetilde{\mathcal{L}_{B}}\left(\theta_{i}^{a}\right). \nonumber
\end{align}
where $\rho_1 >0$ is the perturbed radius, and $\eta >0$ is the learning rate. 

Using the first order Taylor expansion, we have
\begin{align}
 & \nabla_{\theta_{i}}\widetilde{\mathcal{L}_{B}}\left(\theta_{i}^{a}\right)=\nabla_{\theta_{i}}\left[\widetilde{\mathcal{L}_{B}}\left(\theta_{i}+\rho_{1}\frac{\nabla_{\theta_{i}}\widetilde{\mathcal{L}_{B}}\left(\theta_{i}\right)}{\Vert\nabla_{\theta_{i}}\widetilde{\mathcal{L}_{B}}\left(\theta_{i}\right)\Vert}\right)\right]\nonumber \\
\approx & \nabla_{\theta_{i}}\left[\widetilde{\mathcal{L}_{B}}\left(\theta_{i}\right)+\rho_{1} \nabla_{\theta_{i}}\widetilde{\mathcal{L}_{B}}\left(\theta_{i}\right)\cdot\frac{\nabla_{\theta_{i}}\widetilde{\mathcal{L}_{B}}\left(\theta_{i}\right)}{\Vert\nabla_{\theta_{i}}\widetilde{\mathcal{L}_{B}}\left(\theta_{i}\right)\Vert}\right]\nonumber \\ 
&=  \nabla_{\theta_{i}}\left[\widetilde{\mathcal{L}_{B}}\left(\theta_{i}\right)+\rho_{1}\Vert\nabla_{\theta_{i}}\widetilde{\mathcal{L}_{B}}\left(\theta_{i}\right)\Vert\right],\label{eq:sharpness_1}
\end{align}
where $\cdot$ represents the dot product.

\begin{figure}
    \centering
    \includegraphics[width=0.5\columnwidth]{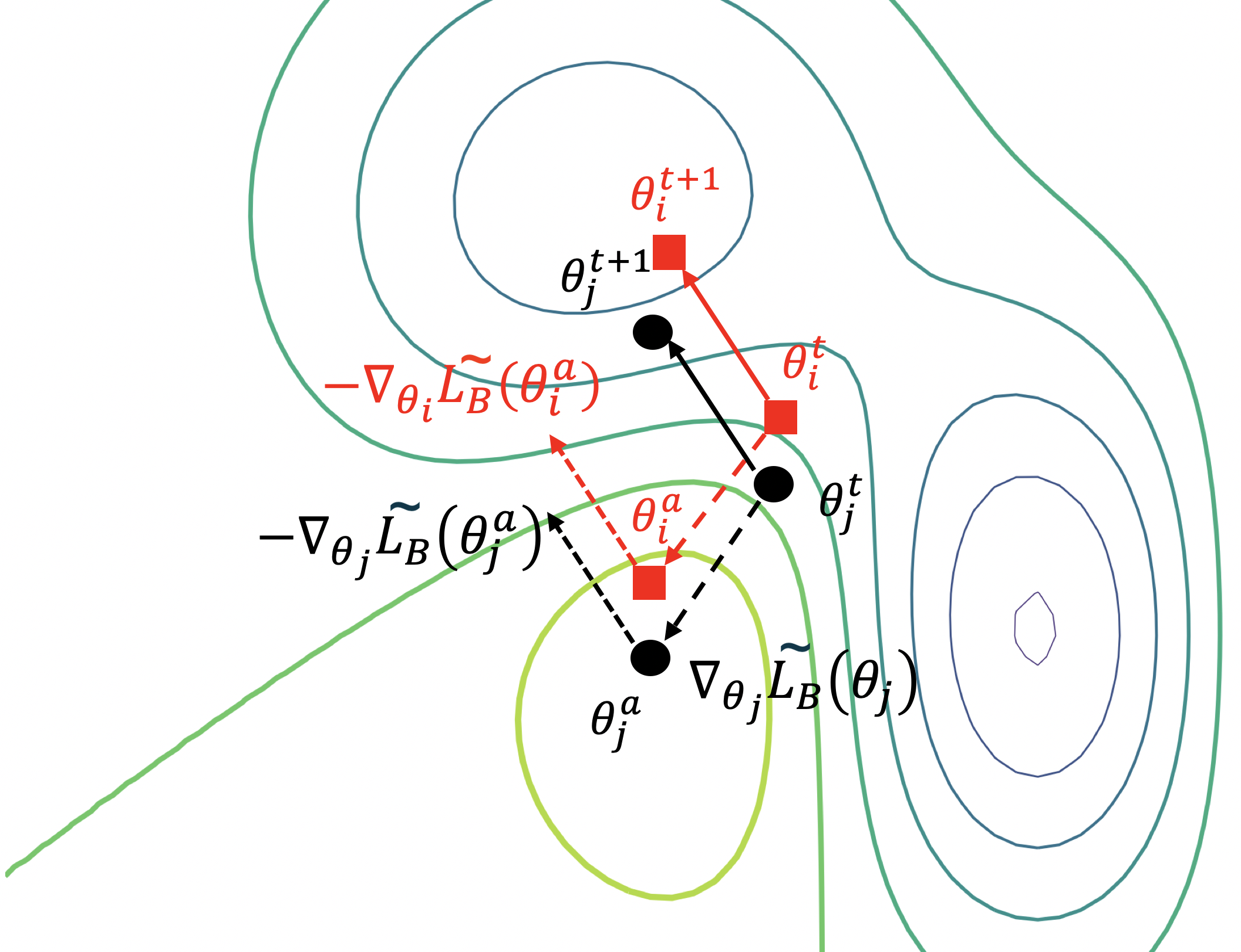}
    \caption{Illustration of the model dynamics under \textbf{sharpness-aware term} on loss landscape. Two base learners $\theta_i$ and $\theta_j$ (represented by the red and black vectors respectively) happen to be initialized closely. At each step, since updated independently yet using the same mini-batch from $\theta_i$ and $\theta_j$, two perturbed models $\theta_i^a$ and $\theta_i^a$ are less diverse, hence two updated models $\theta_i$ and $\theta_j$ are also less diverse and more likely end up at the same low-loss and flat region.} 
    \label{fig:illus-flatness}
\end{figure}

The approximation in \cref{eq:sharpness_1} indicates that since we follow the negative gradient $-\nabla_{\theta_{i}}\widetilde{\mathcal{L}_{B}}\left(\theta_{i}^{a}\right)$ when updating the current model $\theta_i$, 
the new model tends to decrease both the loss $\widetilde{\mathcal{L}_{B}}\left(\theta_{i}\right)$ and the gradient norm $\Vert\nabla_{\theta_{i}}\widetilde{\mathcal{L}_{B}}\left(\theta_{i}\right)\Vert$, directing the base learners to go into the low-loss and flat regions as expected. In this case, there is a possibility that all the base learners, each with sufficient expressitivity, will converge to areas surrounding the same low-loss region. 
Moreover, the normalized gradients $\frac{\nabla_{\theta_{i}}\widetilde{\mathcal{L}_{B}}\left(\theta_{i}\right)}{\Vert\nabla_{\theta_{i}}\widetilde{\mathcal{L}_{B}}\left(\theta_{i}\right)\Vert}, i=1,\dots,m$ reveals that the perturbed models $\theta_i^a, i=1,\dots,m$ are also less diverse because they are computed using the same mini-batch $B$. 
Our intuition is that whenever we add constraints to the objective function $\widetilde{\mathcal{L}_B}(\theta_i)$ of each individual base learner, \textit{if these constraints are independent and do not interact with other base learners, 
the solution space of each base learner is reduced}. Because each base learner is optimized independently, this eventually leads to less diverse updated models $\theta_i, i=1,\dots,m$. We illustrate this intuition in \cref{fig:illus-flatness}.

\vspace{-2em}
Although we expect minimizing sharpness in the ensemble via $\mathcal{L}_B^{\mathrm{ens}}(\cdot)$ would foster a model dynamics where each learner complements each other to support generalization, the above analysis warns us against an adverse effect where ensemble diversity is reduced, thus depriving us of the desirable synergy. The empirical evidence for this intuition can be found in \Cref{tab:abl-flat}. The question now is \textit{how to strengthen the sharpness-aware learning of each individual base learner that interacts with other base learners to achieve both sharpness and diversity?}

To this end, we propose the following agnostic update
\begin{align}
\theta_{i}^{a} & =\theta_{i}+\rho_{1}\frac{\nabla_{\theta_{i}}\widetilde{\mathcal{L}_{B}}\left(\theta_{i}\right)}{\Vert\nabla_{\theta_{i}}\widetilde{\mathcal{L}_{B}}\left(\theta_{i}\right)\Vert}+\rho_{2}\frac{\nabla_{\theta_{i}}\mathcal{L}_{B}^{div}\left(\theta_{i},\theta_{\neq i}\right)}{\Vert\nabla_{\theta_{i}}\mathcal{L}_{B}^{div}\left(\theta_{i},\theta_{\neq i}\right)\Vert},\label{eq:div_SAM_update}\\
\theta_{i} & =\theta_{i}-\eta\nabla_{\theta_{i}}\widetilde{\mathcal{L}_{B}}\left(\theta_{i}^{a}\right),\nonumber 
\end{align}
where $\theta_{\neq i}$ specifies the set of models excluding $\theta_i$ and the $i$-th divergence loss is defined as
\begin{align} \label{eq:l_div}
    \mathcal{L}_{B}^{div}\left(\theta_{i},\theta_{\neq i}\right)= 
    \frac{1}{|B|}\sum_{x\in B, j \neq i}KL\left(\sigma\left(h_{\theta_{i}}^{i}\left(x\right)/\tau\right),\sigma\left(h_{\theta_{j}}^{j}\left(x\right)/\tau\right)\right),
\end{align}
where $h^k_{\theta_k}$ returns \textit{non-targeted logits} (i.e., excluding the logit value of the ground-truth class) of the $k$-th base learner, $\sigma$ is the softmax function, $\tau >0$ is the temperature variable, $\rho_2$ is another perturbed radius, and KL specifies the Kullback-Leibler divergence. 
In practice, we choose $\rho_2=\rho_1$ for simplicity and $\tau < 1$ to favor the distance on dominating modes on each base learner. 

It is worth noting that \cref{eq:l_div} only considers the logits of the \textit{non-targeted} labels for diversifying the base learners, to avoid interfering with their performance on predicting ground-truth labels.
To inspect the agnostic behavior of the second gradient w.r.t the perturbed models $\theta_i^a$, we again use the first-order Taylor expansion

{\small{}
\begin{align}
\nabla_{\theta_{i}}\mathcal{L}_{B}\left(\theta_{i}^{a}\right)= & \nabla_{\theta_{i}}\Biggl[\mathcal{L}_{B}\Biggl(\theta_{i}+\rho_{1}\frac{\nabla_{\theta_{i}}\widetilde{\mathcal{L}_{B}}\left(\theta_{i}\right)}{\Vert\nabla_{\theta_{i}}\widetilde{\mathcal{L}_{B}}\left(\theta_{i}\right)\Vert}\nonumber \\
&+ \rho_{2}\frac{\nabla_{\theta_{i}}\mathcal{L}_{B}^{div}\left(\theta_{i},\theta_{\neq i}\right)}{\Vert\nabla_{\theta_{i}}\mathcal{L}_{B}^{div}\left(\theta_{i},\theta_{\neq i}\right)\Vert}\Biggr]\Biggr)\nonumber \\
\approx & \nabla_{\theta_{i}}\Biggl[\widetilde{\mathcal{L}_{B}}\left(\theta_{i}\right) + \rho_{1}\nabla_{\theta_{i}}\widetilde{\mathcal{L}_{B}}\left(\theta_{i}\right)\cdot\frac{\nabla_{\theta_{i}}\widetilde{\mathcal{L}_{B}}\left(\theta_{i}\right)}{\Vert\nabla_{\theta_{i}}\widetilde{\mathcal{L}_{B}}\left(\theta_{i}\right)\Vert}\nonumber \\
&+ \rho_{2}\nabla_{\theta_{i}}\widetilde{\mathcal{L}_{B}}\left(\theta_{i}\right)\cdot\frac{\nabla_{\theta_{i}}\mathcal{L}_{B}^{div}\left(\theta_{i},\theta_{\neq i}\right)}{\Vert\nabla_{\theta_{i}}\mathcal{L}_{B}^{div}\left(\theta_{i},\theta_{\neq i}\right)\Vert}\Biggr]\nonumber \\
= & \nabla_{\theta_{i}}\Biggl[\widetilde{\mathcal{L}_{B}}\left(\theta_{i}\right)+\rho_{1}\Vert\nabla_{\theta_{i}}\widetilde{\mathcal{L}_{B}}\left(\theta_{i}\right)\Vert \nonumber \\
&-\rho_{2}\frac{-\nabla_{\theta_{i}}\widetilde{\mathcal{L}_{B}}\left(\theta_{i}\right)\cdot\nabla_{\theta_{i}}\mathcal{L}_{B}^{div}\left(\theta_{i},\theta_{\neq i}\right)}{\Vert\nabla_{\theta_{i}}\mathcal{L}_{B}^{div}\left(\theta_{i},\theta_{\neq i}\right)\Vert}\Biggr].\label{eq:div_sharpness}
\end{align}
}

\begin{figure}
    \centering
    \includegraphics[width=0.5\columnwidth]{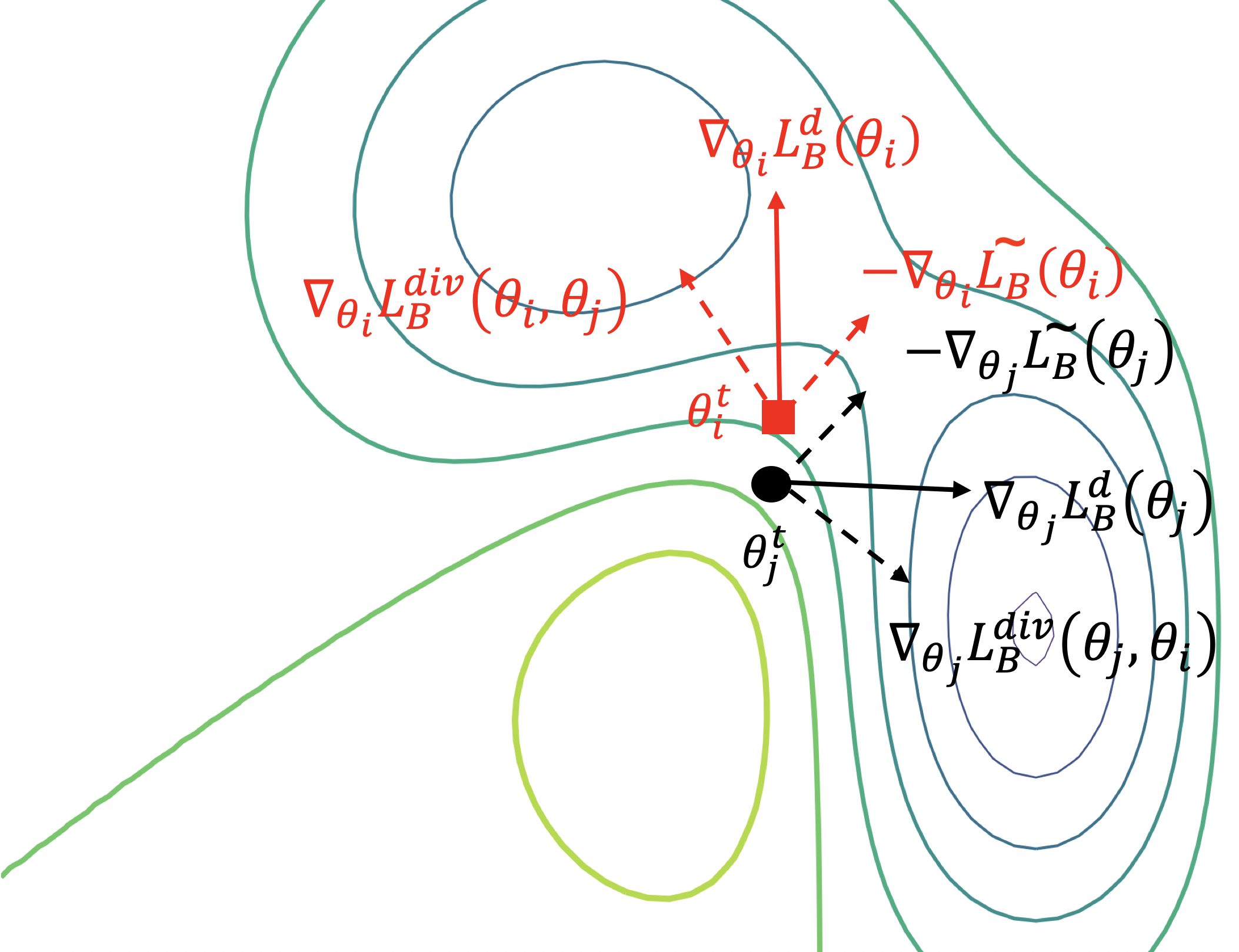}
    \caption{Illustration of the model dynamics under \textbf{diversity-aware term}. Given two base learners $\theta_i$ and $\theta_j$ (represented by the red and black vectors respectively), the gradients $-\nabla_{\theta_i} \widetilde{\mathcal{L}_B}(\theta_i)$ and $-\nabla_{\theta_i} \widetilde{\mathcal{L}_B}(\theta_i)$ navigate the models towards their low-loss (also flat) regions. Moreover, the two gradients $\nabla_{\theta_i} \mathcal{L}_B^{div}(\theta_i, \theta_{\neq i})$ and $\nabla_{\theta_j} \mathcal{L}_B^{div}(\theta_j, \theta_{\neq j})$ encourage the models to move divergently. As discussed, our update strategy forces the two gradients $-\nabla_{\theta_i} \widetilde{\mathcal{L}_B}(\theta_i)$ and $\nabla_{\theta_i} \mathcal{L}_B^{div}(\theta_i, \theta_{\neq i})$ to be more congruent. As the result, two models are divergently oriented to two non-overlapping low-loss and flat regions. This behavior is imposed similarly for the other pair w.r.t. the model $\theta_j$, altogether enhancing the ensemble diversity.} 
    \label{fig:illus-diverse}
\end{figure}

In \cref{eq:div_sharpness}, the first two terms lead the base learners to go to their low-loss and flat regions as discussed before. 
We then analyze the agnostic behavior of the third term. According to the update formula of $\theta_i$ in \cref{eq:div_SAM_update}, 
we follow the positive direction of $\nabla_{\theta_{i}} \mathcal{L}_{\mathcal{B}}^{d} = \nabla_{\theta_{i}}\left[\frac{-\nabla_{\theta_{i}}\widetilde{\mathcal{L}_{B}}\left(\theta_{i}\right)\cdot\nabla_{\theta_{i}}\mathcal{L}_{B}^{div}\left(\theta_{i},\theta_{\neq i}\right)}{\Vert\nabla_{\theta_{i}}\mathcal{L}_{B}^{div}\left(\theta_{i},\theta_{\neq i}\right)\Vert}\right]$, 
further implying that the updated base learner networks aim to maximize $\frac{-\nabla_{\theta_{i}}\widetilde{\mathcal{L}_{B}}\left(\theta_{i}\right)\cdot\nabla_{\theta_{i}}\mathcal{L}_{B}^{div}\left(\theta_{i},\theta_{\neq i}\right)}{\Vert\nabla_{\theta_{i}}\mathcal{L}_{B}^{div}\left(\theta_{i},\theta_{\neq i}\right)\Vert}$.  
Therefore, the low-loss direction $-\nabla_{\theta_{i}}\widetilde{\mathcal{L}_{B}}\left(\theta_{i}\right)$ becomes more congruent with $\frac{\nabla_{\theta_{i}}\mathcal{L}_{B}^{div}\left(\theta_{i},\theta_{\neq i}\right)}{\Vert\nabla_{\theta_{i}}\mathcal{L}_{B}^{div}\left(\theta_{i},\theta_{\neq i}\right)\Vert}$, 
meaning that the base learners tend to diverge while moving along the low-loss and flat directions. \cref{fig:illus-diverse} visualizes our intuition.

\section{Experiments}
We evaluate our methods on the classification tasks on CIFAR10/100 and Tiny-Imagenet. 
We experiment with homogeneous ensembles wherein all base learners has the same model architecture, i.e., R18x3 is an ensemble which consists of three ResNet18 models. We also experiment with heterogeneous ensemble, i.e., RME is an ensemble which consists of ResNet18, MobileNet and EfficientNet models.
The configuration shared between our method and the baselines involves model training for $200$ epochs using SGD optimizer with weight decay of $0.005$. We follow the standard data pre-processing schemes that consists of zero-padding with 4 pixels on each side, random crop, horizon flip and normalization. 
The ensemble prediction has been aggregated by averaging the softmax predictions of all base classifiers.\footnote{Our code is anonymously published at \url{https://anonymous.4open.science/r/DASH/}.} In all tables, bold/underline indicates the best/second-best method. $\uparrow$,$\downarrow$ respectively indicates higher/lower performance is better. We provide our algorithm and more ablation studies in the supplementary materials.

\subsection{Baselines}
This work focuses on improving generalization of ensembles. We compare our method against top ensemble methods with high predictive accuracies across literature: Deep ensembles ~\cite{lakshminarayanan2017simple}, Snapshot ensembles ~\cite{huang2017snapshot}, Fast Geometric Ensemble (FGE) ~\cite{garipov2018loss}, sparse ensembles EDST and DST ~\cite{liudeep}. 
We also deploy SGD and SAM ~\cite{foret2021sharpnessaware} as different optimizers to train an ensemble model and consider as two additional baselines. 

\subsection{Metrics}
We use Ensemble accuracy (Acc) as the primary metric used to measure the generalization of an ensemble learning method. To evaluate the uncertainty capability of a model, we use the standard metrics: Negative Log-Likelihood (NLL), Brier score, and Expected Calibration Error (ECE), which are widely used in the literature. We also report calibrated uncertainty estimation (UE) metrics, such as Cal-NLL, Cal-Brier, and Cal-AAC, at the optimal temperature to avoid measuring calibration error that can be eliminated by simple temperature scaling, as suggested in ~\cite{ashukhapitfalls}.
To measure ensemble diversity, we use Disagreement (D) of predictions ~\cite{kuncheva2003measures} and Log of Determinant (LD) of a matrix consisting of non-target predictions of base classifiers, as proposed in ~\cite{pang2019improving}. The LD metric provides an elegant geometric interpretation of ensemble diversity, which is better than the simple disagreement metric.

\subsection{Evaluation of Predictive Performance}

The results presented in Table \ref{tab:main-acc}  demonstrate the effectiveness of our proposed method, DASH, in improving the generalization ability of ensemble methods. Across all datasets and architectures, DASH consistently and significantly outperformed all baselines. For example, when compared to SGD with R18x3 architecture, DASH achieved substantial improvement gaps of $1.5\%, 3.3\%$, and $7.6\%$ on the CIFAR10, CIFAR100, and Tiny-ImageNet datasets, respectively. When compared to Deep Ensemble, DASH achieved improvement gaps of $3.0\%, 6.8\%$, and $4.0\%$, respectively, on these same datasets.
Our results also provide evidence that seeking more flat classifiers can bring significant benefits to ensemble learning. SAM achieves improvements over SGD or Deep Ensemble, but DASH achieved even greater improvements. Specifically, on the CIFAR100 dataset, DASH outperformed SAM by $3.1\%, 2.1\%$, and $2.3\%$ with R10x5, R18x3, and RME architectures, respectively, while that improvement on the Tiny-ImageNet dataset was $3.8\%$. This improvement indicates the benefits of effectively collaborating between flatness and diversity seeking objectives in deep ensembles. 

Unlike Fast Geometric, Snapshot, or EDST methods, which are limited to homogeneous ensemble settings, DASH is a general method capable of improving ensemble performance even when ensembling different architectures. This is evidenced by the larger improvement gaps over SAM on the RME architecture (i.e., $1.4\%$ improvement on the CIFAR10 dataset) compared to the R18x3 architecture (i.e., $0.9\%$ improvement on the same dataset). These results demonstrate the versatility and effectiveness of DASH in improving the generalization ability of deep ensembles across diverse architectures and datasets.

\begin{table}[hbt!]
    \begin{centering}
    \caption{Ensemble \textbf{accuracy} (\%) on the CIFAR10/100 and Tiny-ImageNet datasets. R10x5 indicates an ensemble of five
    ResNet10 models. R18x3 indicates an ensemble of three ResNet18 models. RME indicates an ensemble of ResNet18, MobileNet and EfficientNet, respectively. 
    }
    \label{tab:main-acc}
    \begin{tabular}{lccccccccc}
    \toprule
    & \multicolumn{3}{c}{CIFAR10} &  & \multicolumn{3}{c}{CIFAR100} &  & Tiny-ImageNet
     \tabularnewline
    \cline{2-4}\cline{3-4} \cline{4-4} \cline{6-8} \cline{7-8} \cline{8-8} \cline{10-10}
    \noalign{\vspace{0.5ex}} 
    Accuracy $\uparrow$  & R10x5 & R18x3 & RME &  & R10x5 & R18x3 & RME &  & R18x3\tabularnewline
    \midrule
    Deep Ensemble & 92.7 & 93.7 & 89.0 &  & 73.7 & 75.4 & 62.7 &  & 65.9\tabularnewline
    Fast Geometric & 92.5 & 93.3 & - &  & 63.2 & 72.3 & - &  & 61.8\tabularnewline
    Snapshot & 93.6 & 94.8 & - &  & 72.8 & 75.7 & - &  & 62.2\tabularnewline
    EDST & 92.0 & 92.8 & - &  & 68.4 & 69.6 & - &  & 62.3\tabularnewline
    DST & 93.2 & 94.7 & 93.4 &  & 70.8 & 70.4 & 71.7 &  & 61.9\tabularnewline
    SGD & 95.1 & 95.2 & 92.6 &  & 75.9 & 78.9 & 72.6 &  & 62.3\tabularnewline
    SAM & \underline{95.4} & \underline{95.8} & \underline{93.8} &  & \underline{77.7} & \underline{80.1} & \underline{76.4} &  & \underline{66.1}\tabularnewline
    DASH (Ours) & \textbf{95.7} & \textbf{96.7} & \textbf{95.2} &  & \textbf{80.8} & \textbf{82.2} & \textbf{78.7} &  & \textbf{69.9}\tabularnewline
    \bottomrule
    \end{tabular}
    \par\end{centering}
    \centering{}
\end{table}
\begin{table}[hbt!]
    \caption{Evaluation of Uncertainty Estimation (UE). \textbf{Calibrated-Brier
    score} is chosen as the representative UE metric reported
    in this table. Evaluation on all \textbf{six} UE metrics for CIFAR10/100
    can be found in the supplementary material. Overall, our method achieves better calibration than baselines on several metrics, especially in the heterogeneous ensemble setting.}
    \label{tab:cal-brier}

    \begin{centering}
    \par\end{centering}
    \centering{}
    \begin{tabular}{lccccccccc}
    \toprule
     & \multicolumn{3}{c}{CIFAR10} &  & \multicolumn{3}{c}{CIFAR100} &  & Tiny-ImageNet\tabularnewline
    \cline{2-4} \cline{3-4} \cline{4-4} \cline{6-8} \cline{7-8} \cline{8-8} \cline{10-10}
    \noalign{\vspace{0.5ex}} 
    Cal-Brier $\downarrow$ & R10x5 & R18x3 & RME &  & R10x5 & R18x3 & RME &  & R18x3\tabularnewline
    \midrule
    Deep Ensemble & 0.091 & 0.079 & 0.153 &  & 0.329 & 0.308 & 0.433 &  & \underline{0.453}\tabularnewline
    Fast Geometric & 0.251 & 0.087 & - &  & 0.606 & 0.344 & - &  & 0.499\tabularnewline
    Snapshot & 0.083 & 0.071 & - &  & 0.338 & 0.311 & - &  & 0.501\tabularnewline
    EDST & 0.122 & 0.113 & - &  & 0.427 & 0.412 & - &  & 0.495\tabularnewline
    DST & 0.102 & 0.083 & 0.102 &  & 0.396 & 0.405 & 0.393 &  & 0.500\tabularnewline
    SGD & 0.078 & 0.076 & 0.113 &  & 0.346 & 0.304 & 0.403 &  & 0.518\tabularnewline
    SAM & \underline{0.073} & \underline{0.067} & \underline{0.094} &  & \underline{0.321} & \underline{0.285} & \underline{0.347} &  & 0.469\tabularnewline
    DASH (Ours) & \textbf{0.067} & \textbf{0.056} & \textbf{0.075} &  & \textbf{0.267} & \textbf{0.255} & \textbf{0.298} &  & \textbf{0.407}\tabularnewline
    \bottomrule
    \end{tabular}
\end{table}

\subsection{Evaluation of Uncertainty Estimation}

Although improving uncertainty estimation is not the primary focus of our method, in this section we still would like to investigate the effectiveness of our method on this aspect by measuring six UE metrics across all experimental settings. 
We present the results of our evaluation in Table \ref{tab:cal-brier}, where we compare the uncertainty estimation capacity of our method with various baselines using the Calibrated-Brier score as the representative metric. Our method consistently achieves the best performance over all baselines across all experimental settings. For instance, on the CIFAR10 dataset with the R10x5 setting, our method obtains a score of 0.067, a relative improvement of 26\% over the Deep Ensemble method. Similarly, across all settings, our method achieves a relative improvement of $26\%, 29\%, 51\%, 18\%, 17\%, 31\%$, and $10\%$ over the Deep Ensemble method.
Furthermore, in Table \ref{tab:ue-tiny-r18x3}, we evaluate the performance of our method on all six UE metrics on the Tiny-ImageNet dataset. In this setting, our method achieves the best performance on five UE metrics, except for the ECE metric. 
Compared to the Deep Ensemble method, our method obtains a relative improvement of $10\%, 3\%$, and $14\%$ on the Cal-Brier, Cal-ACC, and Cal-NLL metrics, respectively.
In conclusion, our method shows promising results in improving uncertainty estimation, as demonstrated by its superior performance in various UE metrics.

\begin{table}
\caption{Evaluation of Uncertainty Estimation (UE) across six standard UE metrics on the Tiny-ImageNet dataset with R18x3.}
\label{tab:ue-tiny-r18x3}
\centering{}
\begin{tabular}{lcccccc}
\toprule
 & NLL $\downarrow$ & Brier $\downarrow$ & ECE $\downarrow$ & Cal-Brier $\downarrow$ & Cal-AAC $\downarrow$ & Cal-NLL $\downarrow$\tabularnewline
\midrule
Deep Ensemble & 1.400 & 0.452 & \textbf{0.110} & 0.453 & 0.210 & 1.413\tabularnewline
Fast Geometric & 1.548 & 0.501 & 0.116 & 0.499 & 0.239 & 1.544\tabularnewline
Snapshot & 1.643 & 0.505 & 0.118 & 0.501 & 0.237 & 1.599\tabularnewline
EDST & 1.581 & 0.496 & 0.115 & 0.495 & 0.235 & 1.548\tabularnewline
DST & 1.525 & 0.499 & \textbf{0.110} & 0.500 & 0.239 & 1.536\tabularnewline
SGD & 1.999 & 0.601 & 0.283 & 0.518 & 0.272 & 1.737\tabularnewline
SAM & 1.791 & 0.563 & 0.297 & 0.469 & 0.242 & 1.484\tabularnewline
DASH (Ours) & \textbf{1.379} & \textbf{0.447} & 0.184 & \textbf{0.407} & \textbf{0.204} & \textbf{1.213}\tabularnewline
\bottomrule
\end{tabular}
\end{table}
\raggedbottom

\subsection{Evaluation on Adversarial Robustness}

In this section, our goal is to evaluate the adversarial robustness of our proposed method against adversarial attacks. To achieve this, we conducted experiments on the CIFAR10 dataset using the R18x3 architecture and employed the PGD attack ~\cite{madrytowards}, which is considered the standard adversarial attack for evaluating robustness. Specifically, we set the number of attack steps to $k=10$, step size to $\eta=1/255$, and varied the change in perturbation size $\epsilon$ from $1/255$ to $6/255$.

While it is widely recognized in the Adversarial Machine Learning literature that strong attacks are required to truly challenge defense methods (i.e., PGD attack with more than 200 attack steps with a perturbation size of $\epsilon=8/255$), we chose a weaker attack for our experiments. This decision was based on the fact that all methods we evaluated were not specifically designed to enhance adversarial robustness, and therefore may not perform well against a stronger attack.

\begin{figure*}
\centering
\begin{subfigure}{.5\textwidth}
  \centering
  \includegraphics[width=1.0\textwidth]{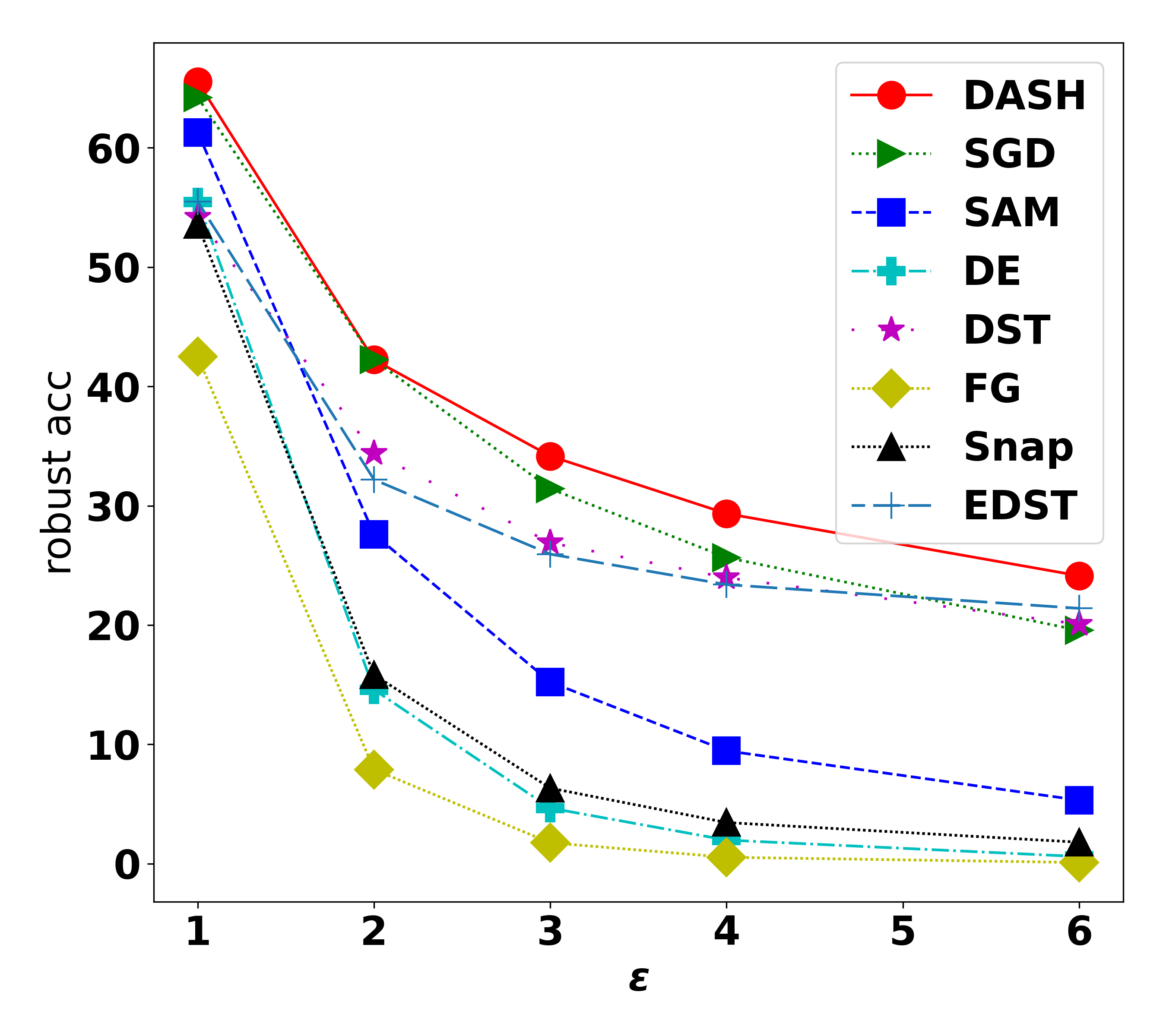}
  \caption{R18x3}
  \label{fig:adv_cifar10_r18x3}
\end{subfigure}%
\begin{subfigure}{.5\textwidth}
  \centering
  \includegraphics[width=1.0\textwidth]{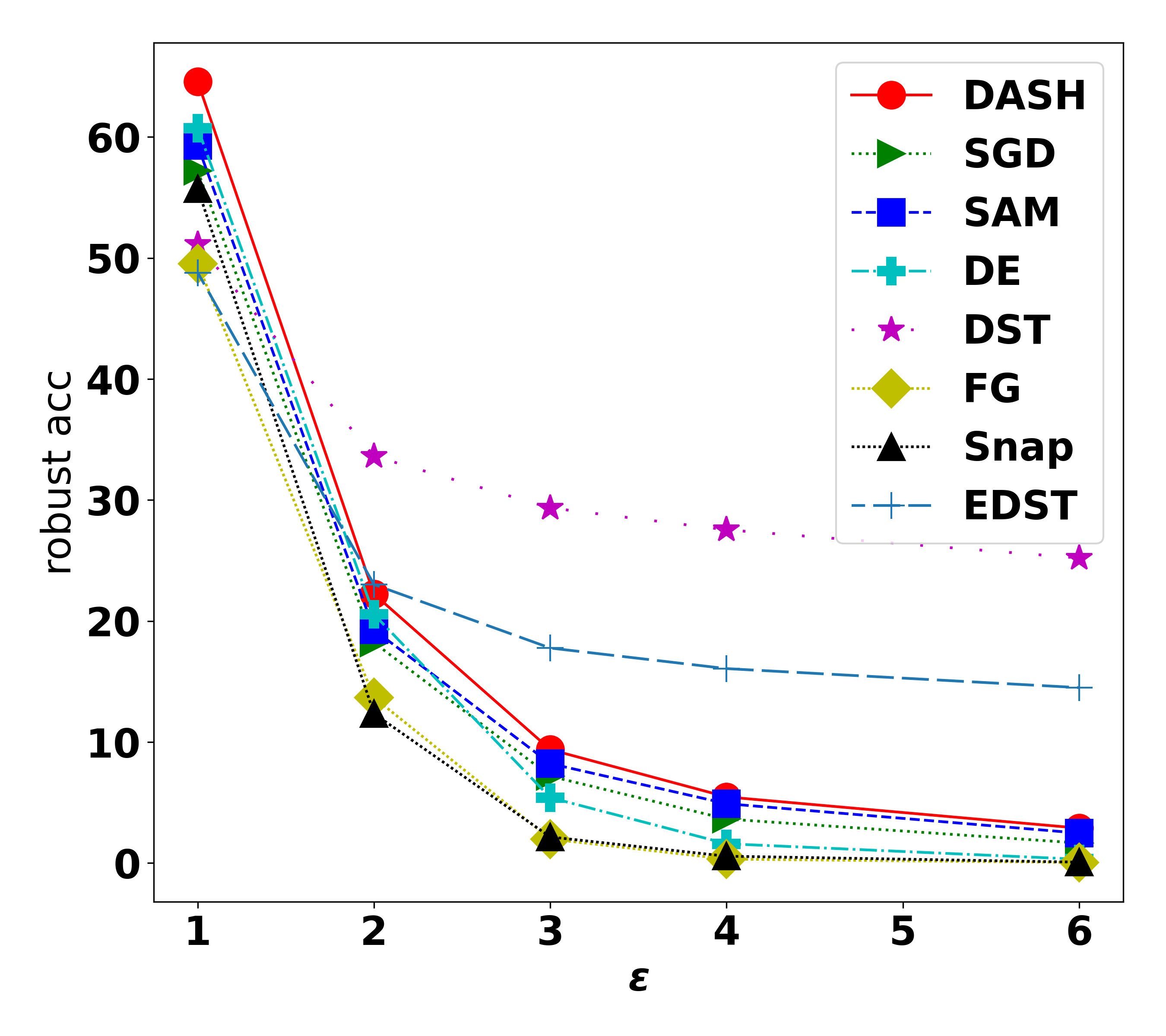}
  \caption{R10x5}
  \label{fig:adv_cifar10_r10x5}
\end{subfigure}

\caption{Evaluation on Adversarial Robustness. The x-axis denotes the perturbation size $\epsilon$ (*255). }
\label{fig:adv_cifar10}
\end{figure*}

It can be seen from \cref{fig:adv_cifar10_r18x3} that our DASH achieves better adversarial robustness than all baselines on the R18x3 architecture. More specifically, our method consistently outperforms SGD by around 3\% across different levels of $\epsilon$. While there is a huge drop of adversarial robustness on SAM when the attack becomes stronger (i.e., 61.28\% with $\epsilon=1/255$ and 27.61\% with $\epsilon=2/255$), our method is more robust with a smaller drop (i.e., 65.53\% with $\epsilon=1/255$ and 42.23\% with $\epsilon=2/255$). 
On the R10x5 architecture, our method still outperforms SGD and SAM across all levels of attack strength. However, it can be observed that our DASH achieves a lower performance than DST and EDST methods if the perturbation size $\epsilon \geq 2/255$ as shown in \cref{fig:adv_cifar10_r10x5}. While our method does not specifically target improving adversarial robustness, the superior performance we achieve on the R18x3 architecture suggests that our principle of considering sharpness-aware and diverse-aware mechanisms could be a promising direction for addressing this issue. 
 
\section{Ablation studies}
\subsection{Hyper-parameter sensitivity}
Table \ref{tab:tune-gamma} reports the effect of the hyper-parameter $\gamma$ on the performance of our method by tuning it over the range of $[0, 1]$. Recall that $\gamma=0.1$ means that we prioritize seeking flatness on all individual base classifiers over the entire ensemble model, while $\gamma=1$ means that we only seek flatness on the entire aggregated ensemble model only.
We conduct the experiment on the CIFAR100 dataset with R10x5 architecture and report results on Table \ref{tab:tune-gamma}.
It can be seen that our method achieves the best performance in both generalization and uncertainty estimation aspects when $\gamma=0.1$ and there is a significant drop of $1.8\%$ in accuracy when $\gamma=1$. 
In our experiments, we set $\gamma=0.1$ as the default setting.

\begin{table}[hbt!]
\caption{Ensemble performance under various the trade-off parameters $\gamma$ on the CIFAR100 dataset with R10x5 architecture.}
\label{tab:tune-gamma}
\begin{centering}
\begin{tabular}{lcccc}
\toprule
 & Accuracy  $\uparrow$ & NLL $\downarrow$ & Brier $\downarrow$ & ECE $\downarrow$\tabularnewline
\midrule 
$\gamma=0.1$ & \textbf{80.84} & \textbf{0.86} & \textbf{0.32} & \textbf{0.18}\tabularnewline
$\gamma=0.2$ & 80.48 & 0.97 & 0.35 & 0.23\tabularnewline
$\gamma=0.5$ & 80.42 & 0.95 & 0.34 & 0.22\tabularnewline
$\gamma=0.8$ & 79.81 & 1.08 & 0.38 & 0.29\tabularnewline
$\gamma=1.0$ & 78.86 & 1.12 & 0.40 & 0.28\tabularnewline
\bottomrule
\end{tabular}
\par\end{centering}
\begin{centering}
\par\end{centering}
\centering{}
\end{table}
\raggedbottom

\subsection{Contribution of the diverse-aware agnostic constraint}

In this section, our objective is to assess the impact of each component by comparing the performance of two variants: DASH and $\text{DASH}^F$, where the latter is our method with flat seeking mode only. We run the experiments on the CIFAR10 and CIFAR100 datasets with RME architecture, and the results are presented in Table \ref{tab:abl-flat}.
We observed that $\text{DASH}^F$ outperforms the standard SGD method by a substantial margin when using the flat seeking mode only. The performance improvement is remarkable, with a gap of $1.72\%$ and $3.73\% $on the CIFAR10 and CIFAR100 datasets, respectively. This enhancement can be attributed to the improvement of each single base classifier. The ensemble can achieve better generalization performance by combining these classifiers. In particular, the average accuracy of all base classifiers with $\text{DASH}^F$ is $93.21\%$, which is $5.07\%$ higher than that achieved with the SGD method.

However, in terms of ensemble diversity, measured by the Log-Determinant metric, $\text{DASH}^F$'s base classifiers are less diverse than those of SGD. Specifically, on the same CIFAR100 dataset, SGD obtains a LD score of $-16.88$, while that of $\text{DASH}^F$ is only $-19.47$, which is a $15.3\%$ relatively lower. The lower LD score indicates that the predictions of the base classifiers on $\text{DASH}^F$ have a higher similarity than those on SGD. Consequently, in some hard negative samples, the predictions of all base classifiers fall into similar incorrect patterns, and the final ensemble prediction becomes incorrect. 
On the other hand, when comparing between DASH and $\text{DASH}^F$, it can be observed that, DASH obtains a higher LD score in both datasets, while also improves the average performance of the base classifiers. As consequent, DASH improves over $\text{DASH}^F$ by $0.84\%$ to $2.44\%$ on the CIFAR10 and CIFAR100, respectively. 

\begin{table}[hbt!]
\caption{Ablation study on the contribution of each component on the CIFAR10
(C10) and CIFAR100 (C100) datasets with RME architecture. $\text{DASH}^{F}$
represents our method with flat seeking mode only.\label{tab:abl-flat}}
\begin{centering}
\begin{tabular}{llcccc}
\toprule
 &  & Accuracy  $\uparrow$ & LD  $\uparrow$ & D  $\uparrow$ & Avg. Accuracy   $\uparrow$\tabularnewline
\midrule 
\multirow{3}{*}{C10} & SGD & 92.61 & -24.7 & \textbf{0.149} & 88.14\tabularnewline
 & $\text{DASH}^{F}$ & 94.33 & -25.8 & 0.034 & 93.21\tabularnewline
 & $\text{DASH}$ & \textbf{95.17} & \textbf{-23.3} & 0.068 & \textbf{93.41}\tabularnewline
\midrule
\multirow{3}{*}{C100} & SGD & 72.55 & \textbf{-16.88} & \textbf{0.853} & 38.09\tabularnewline
 & $\text{DASH}^{F}$ & 76.28 & -19.47 & 0.123 & 73.38\tabularnewline
 & $\text{DASH}$ & \textbf{78.72} & -18.92 & 0.237 & \textbf{74.69}\tabularnewline
\bottomrule
\end{tabular}
\par\end{centering}
\centering{}
\end{table}

\section{Conclusion}
We developed \textbf{DASH} Ensemble - a learning algorithm that optimizes for deep ensembles of diverse and flat minimizers. Our method begins with a theoretical development to minimize sharpness-aware upper bound for the general loss of the ensemble, followed by a novel addition of an agnostic term to promote divergence among base classifiers. Our experimental results support the effectiveness of the agnostic term in introducing diversity in individual predictions, which ultimately leads to an improvement in generalization performance. This work has demonstrated the potential of integrating sharpness-aware minimization technique into the ensemble learning paradigm. We thus hope to motivate future works to exploit such a connection to develop more powerful and efficient ensemble models.

%
%
\bibliographystyle{splncs04}
\bibliography{dash}

\begin{thebibliography}{10}
\providecommand{\url}[1]{\texttt{#1}}
\providecommand{\urlprefix}{URL }
\providecommand{\doi}[1]{https://doi.org/#1}

\bibitem{abbas2022sharp}
Abbas, M., Xiao, Q., Chen, L., Chen, P.Y., Chen, T.: Sharp-maml: Sharpness-aware model-agnostic meta learning. arXiv preprint arXiv:2206.03996  (2022)

\bibitem{allen2022towards}
Allen-Zhu, Z., Li, Y.: Towards understanding ensemble, knowledge distillation and self-distillation in deep learning. In: The Eleventh International Conference on Learning Representations (2022)

\bibitem{ashukhapitfalls}
Ashukha, A., Lyzhov, A., Molchanov, D., Vetrov, D.: Pitfalls of in-domain uncertainty estimation and ensembling in deep learning. In: International Conference on Learning Representations (2020)

\bibitem{bahri-etal-2022-sharpness}
Bahri, D., Mobahi, H., Tay, Y.: Sharpness-aware minimization improves language model generalization. In: Proceedings of the 60th Annual Meeting of the Association for Computational Linguistics. Association for Computational Linguistics, Dublin, Ireland (May 2022). \doi{10.18653/v1/2022.acl-long.508}, \url{https://aclanthology.org/2022.acl-long.508}

\bibitem{breiman1996bagging}
Breiman, L.: Bagging predictors. Machine learning  \textbf{24} (1996)

\bibitem{breiman1996bias}
Breiman, L.: Bias, variance, and arcing classifiers. Tech. rep., Tech. Rep. 460, Statistics Department, University of California, Berkeley~… (1996)

\bibitem{cha2021swad}
Cha, J., Chun, S., Lee, K., Cho, H.C., Park, S., Lee, Y., Park, S.: Swad: Domain generalization by seeking flat minima. Advances in Neural Information Processing Systems  \textbf{34} (2021)

\bibitem{Chaudhari2017EntropySGDBG}
Chaudhari, P., Choromańska, A., Soatto, S., LeCun, Y., Baldassi, C., Borgs, C., Chayes, J.T., Sagun, L., Zecchina, R.: Entropy-sgd: biasing gradient descent into wide valleys. Journal of Statistical Mechanics: Theory and Experiment  (2019)

\bibitem{chen2021vision}
Chen, X., Hsieh, C.J., Gong, B.: When vision transformers outperform resnets without pre-training or strong data augmentations. arXiv preprint arXiv:2106.01548  (2021)

\bibitem{dietterich2000ensemble}
Dietterich, T.G.: Ensemble methods in machine learning. In: Multiple Classifier Systems: First International Workshop, Proceedings 1. Springer (2000)

\bibitem{dinh2017sharp}
Dinh, L., Pascanu, R., Bengio, S., Bengio, Y.: Sharp minima can generalize for deep nets. In: ICML (2017)

\bibitem{DziugaiteR17}
Dziugaite, G.K., Roy, D.M.: Computing nonvacuous generalization bounds for deep (stochastic) neural networks with many more parameters than training data. In: {UAI}. {AUAI} Press (2017)

\bibitem{evci2020rigging}
Evci, U., Gale, T., Menick, J., Castro, P.S., Elsen, E.: Rigging the lottery: Making all tickets winners. In: International Conference on Machine Learning. PMLR (2020)

\bibitem{evci2022gradient}
Evci, U., Ioannou, Y., Keskin, C., Dauphin, Y.: Gradient flow in sparse neural networks and how lottery tickets win. In: Proceedings of the AAAI Conference on Artificial Intelligence. vol.~36 (2022)

\bibitem{foret2021sharpnessaware}
Foret, P., Kleiner, A., Mobahi, H., Neyshabur, B.: Sharpness-aware minimization for efficiently improving generalization. In: International Conference on Learning Representations (2021), \url{https://openreview.net/forum?id=6Tm1mposlrM}

\bibitem{fort2019emergent}
Fort, S., Ganguli, S.: Emergent properties of the local geometry of neural loss landscapes. arXiv preprint arXiv:1910.05929  (2019)

\bibitem{fort2019deep}
Fort, S., Hu, H., Lakshminarayanan, B.: Deep ensembles: A loss landscape perspective. arXiv preprint arXiv:1912.02757  (2019)

\bibitem{freund1996experiments}
Freund, Y., Schapire, R.E., et~al.: Experiments with a new boosting algorithm. In: ICML. vol.~96 (1996)

\bibitem{friedman2001greedy}
Friedman, J.H.: Greedy function approximation: a gradient boosting machine. Annals of statistics  (2001)

\bibitem{gal2016dropout}
Gal, Y., Ghahramani, Z.: Dropout as a bayesian approximation: Representing model uncertainty in deep learning. In: international conference on machine learning. PMLR (2016)

\bibitem{garipov2018loss}
Garipov, T., Izmailov, P., Podoprikhin, D., Vetrov, D.P., Wilson, A.G.: Loss surfaces, mode connectivity, and fast ensembling of dnns. Advances in neural information processing systems  \textbf{31} (2018)

\bibitem{gustafsson2020evaluating}
Gustafsson, F.K., Danelljan, M., Schon, T.B.: Evaluating scalable bayesian deep learning methods for robust computer vision. In: Proceedings of the IEEE/CVF conference on computer vision and pattern recognition workshops (2020)

\bibitem{hansen1990neural}
Hansen, L.K., Salamon, P.: Neural network ensembles. IEEE transactions on pattern analysis and machine intelligence  \textbf{12}(10) (1990)

\bibitem{havasi2020training}
Havasi, M., Jenatton, R., Fort, S., Liu, J.Z., Snoek, J., Lakshminarayanan, B., Dai, A.M., Tran, D.: Training independent subnetworks for robust prediction. arXiv preprint arXiv:2010.06610  (2020)

\bibitem{DBLP:conf/nips/HochreiterS94}
Hochreiter, S., Schmidhuber, J.: Simplifying neural nets by discovering flat minima. In: {NIPS}. {MIT} Press (1994)

\bibitem{huang2017snapshot}
Huang, G., Li, Y., Pleiss, G., Liu, Z., Hopcroft, J.E., Weinberger, K.Q.: Snapshot ensembles: Train 1, get m for free. arXiv preprint arXiv:1704.00109  (2017)

\bibitem{huang2016deep}
Huang, G., Sun, Y., Liu, Z., Sedra, D., Weinberger, K.Q.: Deep networks with stochastic depth. In: Computer Vision--ECCV 2016: 14th European Conference, Amsterdam, The Netherlands, October 11--14, 2016, Proceedings, Part IV 14. Springer (2016)

\bibitem{Jastrzebski2017ThreeFI}
Jastrzebski, S., Kenton, Z., Arpit, D., Ballas, N., Fischer, A., Bengio, Y., Storkey, A.J.: Three factors influencing minima in sgd. ArXiv  \textbf{abs/1711.04623} (2017)

\bibitem{JiangNMKB20}
Jiang, Y., Neyshabur, B., Mobahi, H., Krishnan, D., Bengio, S.: Fantastic generalization measures and where to find them. In: {ICLR}. OpenReview.net (2020)

\bibitem{DBLP:conf/iclr/KeskarMNST17}
Keskar, N.S., Mudigere, D., Nocedal, J., Smelyanskiy, M., Tang, P.T.P.: On large-batch training for deep learning: Generalization gap and sharp minima. In: {ICLR}. OpenReview.net (2017)

\bibitem{krogh1994neural}
Krogh, A., Vedelsby, J.: Neural network ensembles, cross validation, and active learning. Advances in neural information processing systems  \textbf{7} (1994)

\bibitem{kuncheva2003measures}
Kuncheva, L.I., Whitaker, C.J.: Measures of diversity in classifier ensembles and their relationship with the ensemble accuracy. Machine learning  \textbf{51}(2) (2003)

\bibitem{lakshminarayanan2017simple}
Lakshminarayanan, B., Pritzel, A., Blundell, C.: Simple and scalable predictive uncertainty estimation using deep ensembles. Advances in neural information processing systems  \textbf{30} (2017)

\bibitem{li2018visualizing}
Li, H., Xu, Z., Taylor, G., Studer, C., Goldstein, T.: Visualizing the loss landscape of neural nets. Advances in neural information processing systems  \textbf{31} (2018)

\bibitem{li2012diversity}
Li, N., Yu, Y., Zhou, Z.H.: Diversity regularized ensemble pruning. In: Machine Learning and Knowledge Discovery in Databases: European Conference, ECML PKDD 2012, Bristol, UK, September 24-28, 2012. Proceedings, Part I 23. pp. 330--345. Springer (2012)

\bibitem{liudeep}
Liu, S., Chen, T., Atashgahi, Z., Chen, X., Sokar, G., Mocanu, E., Pechenizkiy, M., Wang, Z., Mocanu, D.C.: Deep ensembling with no overhead for either training or testing: The all-round blessings of dynamic sparsity. In: International Conference on Learning Representations (2022)

\bibitem{liu2021sparse}
Liu, S., Mocanu, D.C., Matavalam, A.R.R., Pei, Y., Pechenizkiy, M.: Sparse evolutionary deep learning with over one million artificial neurons on commodity hardware. Neural Computing and Applications  \textbf{33} (2021)

\bibitem{madrytowards}
Madry, A., Makelov, A., Schmidt, L., Tsipras, D., Vladu, A.: Towards deep learning models resistant to adversarial attacks. In: International Conference on Learning Representations (2017)

\bibitem{neyshabur2017exploring}
Neyshabur, B., Bhojanapalli, S., McAllester, D., Srebro, N.: Exploring generalization in deep learning. Advances in neural information processing systems  \textbf{30} (2017)

\bibitem{ortega2022diversity}
Ortega, L.A., Caba{\~n}as, R., Masegosa, A.: Diversity and generalization in neural network ensembles. In: International Conference on Artificial Intelligence and Statistics. pp. 11720--11743. PMLR (2022)

\bibitem{ovadia2019can}
Ovadia, Y., Fertig, E., Ren, J., Nado, Z., Sculley, D., Nowozin, S., Dillon, J., Lakshminarayanan, B., Snoek, J.: Can you trust your model's uncertainty? evaluating predictive uncertainty under dataset shift. Advances in neural information processing systems  \textbf{32} (2019)

\bibitem{pang2019improving}
Pang, T., Xu, K., Du, C., Chen, N., Zhu, J.: Improving adversarial robustness via promoting ensemble diversity. In: International Conference on Machine Learning. PMLR (2019)

\bibitem{DBLP:conf/iclr/PereyraTCKH17}
Pereyra, G., Tucker, G., Chorowski, J., Kaiser, L., Hinton, G.E.: Regularizing neural networks by penalizing confident output distributions. In: {ICLR} (Workshop). OpenReview.net (2017)

\bibitem{perrone1995networks}
Perrone, M.P., Cooper, L.N.: When networks disagree: Ensemble methods for hybrid neural networks. In: How We Learn; How We Remember: Toward An Understanding Of Brain And Neural Systems: Selected Papers of Leon N Cooper. World Scientific (1995)

\bibitem{PetzkaKASB21}
Petzka, H., Kamp, M., Adilova, L., Sminchisescu, C., Boley, M.: Relative flatness and generalization. In: NeurIPS (2021)

\bibitem{phan2022improving}
Phan, H., Tran, L., Tran, N.N., Ho, N., Phung, D., Le, T.: Improving multi-task learning via seeking task-based flat regions. arXiv preprint arXiv:2211.13723  (2022)

\bibitem{qu2022generalized}
Qu, Z., Li, X., Duan, R., Liu, Y., Tang, B., Lu, Z.: Generalized federated learning via sharpness aware minimization. arXiv preprint arXiv:2206.02618  (2022)

\bibitem{sinha2020diversity}
Sinha, S., Bharadhwaj, H., Goyal, A., Larochelle, H., Garg, A., Shkurti, F.: Diversity inducing information bottleneck in model ensembles. arXiv preprint arXiv:2003.04514  (2020)

\bibitem{srivastava2014dropout}
Srivastava, N., Hinton, G., Krizhevsky, A., Sutskever, I., Salakhutdinov, R.: Dropout: a simple way to prevent neural networks from overfitting. The journal of machine learning research  \textbf{15}(1) (2014)

\bibitem{sun2018structural}
Sun, T., Zhou, Z.H.: Structural diversity for decision tree ensemble learning. Frontiers of Computer Science  \textbf{12},  560--570 (2018)

\bibitem{wan2013regularization}
Wan, L., Zeiler, M., Zhang, S., Le~Cun, Y., Fergus, R.: Regularization of neural networks using dropconnect. In: International conference on machine learning. PMLR (2013)

\bibitem{wei2020implicit}
Wei, C., Kakade, S., Ma, T.: The implicit and explicit regularization effects of dropout. In: International conference on machine learning. PMLR (2020)

\bibitem{wen2020batchensemble}
Wen, Y., Tran, D., Ba, J.: Batchensemble: an alternative approach to efficient ensemble and lifelong learning. arXiv preprint arXiv:2002.06715  (2020)

\bibitem{wenzel2020hyperparameter}
Wenzel, F., Snoek, J., Tran, D., Jenatton, R.: Hyperparameter ensembles for robustness and uncertainty quantification. Advances in Neural Information Processing Systems  \textbf{33} (2020)

\bibitem{yang2020dverge}
Yang, H., Zhang, J., Dong, H., Inkawhich, N., Gardner, A., Touchet, A., Wilkes, W., Berry, H., Li, H.: Dverge: diversifying vulnerabilities for enhanced robust generation of ensembles. Advances in Neural Information Processing Systems  \textbf{33} (2020)

\bibitem{yang2021trs}
Yang, Z., Li, L., Xu, X., Zuo, S., Chen, Q., Zhou, P., Rubinstein, B., Zhang, C., Li, B.: Trs: Transferability reduced ensemble via promoting gradient diversity and model smoothness. Advances in Neural Information Processing Systems  \textbf{34} (2021)

\bibitem{zhang2008rotboost}
Zhang, C.X., Zhang, J.S.: Rotboost: A technique for combining rotation forest and adaboost. Pattern recognition letters  \textbf{29}(10) (2008)

\bibitem{zhang2019your}
Zhang, L., Song, J., Gao, A., Chen, J., Bao, C., Ma, K.: Be your own teacher: Improve the performance of convolutional neural networks via self distillation. In: Proceedings of the IEEE/CVF International Conference on Computer Vision (2019)

\bibitem{zhang2020diversified}
Zhang, S., Liu, M., Yan, J.: The diversified ensemble neural network. Advances in Neural Information Processing Systems  \textbf{33},  16001--16011 (2020)

\bibitem{Zhang2018DeepML}
Zhang, Y., Xiang, T., Hospedales, T.M., Lu, H.: Deep mutual learning. 2018 IEEE/CVF Conference on Computer Vision and Pattern Recognition  (2018)

\end{thebibliography}
\end{document}


\title{Diversity-Aware Agnostic Ensemble of\\ Sharpness Minimizers}

\author{First Author\inst{1}\orcidlink{0000-1111-2222-3333} \and
Second Author\inst{2,3}\orcidlink{1111-2222-3333-4444} \and
Third Author\inst{3}\orcidlink{2222--3333-4444-5555}}

\authorrunning{F.~Author et al.}

\institute{Princeton University, Princeton NJ 08544, USA \and
Springer Heidelberg, Tiergartenstr.~17, 69121 Heidelberg, Germany
\email{lncs@springer.com}\\
\url{http://www.springer.com/gp/computer-science/lncs} \and
ABC Institute, Rupert-Karls-University Heidelberg, Heidelberg, Germany\\
\email{\{abc,lncs\}@uni-heidelberg.de}}

\subtitle{Supplementary Materials}
\maketitle

\section{Proof for Theorem 1} \label{sec:proofs}
\textbf{Theorem 1.} Assume that the loss function $\ell$ is convex
and upper-bounded by $L$.  
With the probability at least $1-\delta$
over the choices of $\mathcal{S}\sim\mathcal{D}^{N}$, for any $0\leq\gamma\leq1$,
we have
\begin{align*}
    \Lc_{\Dc}(\theta) &\leq \gamma \max_{\theta^{\prime}: \|\theta^{\prime} -\theta\|\leq \sqrt{m} \rho} \Lc_{\Sc}(\theta^{\prime}) + \frac{1-\gamma}{m} \Big[\sum_{i=1}^m\max_{\theta^{\prime}_i:\|\theta^{\prime}_i - \theta_i\|\leq \rho} \Lc_{\Sc}(\theta^{\prime}_i)\Big] +  
   \frac{CL}{\sqrt{N}} \times\\ &\qquad \Bigg[m\sqrt{\log\frac{m(N+k)}{\delta} } + \sum_{i=1}^m \sqrt{k\log\Big(1+\frac{\|\theta_i\|^2}{\rho^2}(1+\sqrt{\log(N)}/k)^2\Big)} +\\
    &\qquad \sqrt{km \log\Big(1+ \frac{\sum_{i=1}^m \|\theta_i\|^2}{m\rho^2}\big(1+ \sqrt{\log(N)/(mk)}\big)^2 \Big)} + O(1)\Bigg]
\end{align*}
 where the $\theta_i$ and the loss function $\ell$ satisfying the conditions: for all $\rho >0$, $P_i \sim \Nc(\theta_i,\rho^2 \mathbb{I}_k)$ and $P=\Nc(\theta,\rho^2 \mathbb{I}_{mk})$
 \begin{align*}
    \mathbb{E}_{(x,y)\in \mathcal{D}}\big[\ell( f_{\theta_i}^i(x),y) \big] &\leq  \mathbb{E}_{\theta^{\prime}_i \sim P_i}\mathbb{E}_{(x,y)\in\mathcal{D}} \big[\ell(f_{\theta_i^{\prime}}^{i}(x),y)\big]\\
    \mathbb{E}_{(x,y)\in \mathcal{D}}\big[\ell( f_{\theta}^{\mathrm{ens}}(x),y) \big] &\leq  \mathbb{E}_{\theta \sim P}\mathbb{E}_{(x,y)\in\mathcal{D}} \big[\ell(f_{\theta}^{\mathrm{ens}}(x),y)\big]
 \end{align*}
where k is the number of parameters.

\begin{proof}
We use the PAC-Bayes theory in this proof. In PAC-Bayes theory, $\theta$ could follow a distribution, says $P$, thus we define the expected loss over $\theta$ distributed by $P$ as follows:
\begin{align*}
    \Lc_{\Dc}(\theta,P) &= \E_{\theta\sim P}\big[\ell_{\Dc}(\theta) \big] \\
    \Lc_{\Sc}(\theta,P) &= \E_{\theta\sim P}\big[\ell_{\Sc}(\theta) \big].
\end{align*}
For any distribution $P= \mathcal{N}(\mathbf{0},\sigma_P^2\mathbb{I}_k)$ and $Q=\mathcal{N}(\theta,\sigma^2\mathbb{I}_k)$ over $\theta\in \mathbb{R}^k$, where $P$ is the prior distribution and $Q$ is the posterior distribution, use the PAC-Bayes theorem in \cite{JMLR:v17:15-290}, for all $\beta>0$, with a probability at least $1-\delta$, we have
    \begin{equation}
        \Lc_{\Dc}(\theta,Q) \leq \Lc_{\Sc}(\theta,Q) +\frac{1}{\beta}\Big[\mathsf{KL}(Q\|P) + \log\frac{1}{\delta} + \Psi(\beta,N) \Big],\label{ineq:PAC-Bayes}
    \end{equation}
    where $\Psi$ is defined as
    \begin{align*}
        \Psi(\beta,N) = \log \E_{P}\E_{\Dc^N}\Big[\exp\big\{\beta\big[\Lc_{\Dc}(f_{\theta}) - \Lc_{\Sc}(f_{\theta})\big] \big\} \Big].
    \end{align*}
When the loss function is bounded by $L$, then 
    \begin{align*}
        \Psi(\beta,N) \leq \frac{\beta^2L^2}{8N}.
    \end{align*}
 The task is to minimize the second term of RHS of \eqref{ineq:PAC-Bayes},  we thus choose $\beta =\sqrt{8N} \frac{\mathsf{KL}(Q\|P) + \log\frac{1}{\delta}}{L}$. Then  the second term of RHS of \eqref{ineq:PAC-Bayes} is equal to
\begin{align*}
\sqrt{\frac{\mathsf{KL}(Q\|P) + \log \frac{1}{\delta}}{2N}}\times L.
\end{align*}
The KL divergence between $Q$ and $P$, when they are Gaussian, is given by formula
\begin{align*}
\mathsf{KL}(Q\|P) = \frac{1}{2}\left[ \frac{k\sigma^2 + \|\theta\|^2}{\sigma_P^2} - k + k \log\frac{\sigma_P^2}{\sigma^2}\right].
\end{align*}
For given posterior distribution $Q$ with fixed $\sigma^2$, to minimize the KL term, the $\sigma_P^2$ should be equal to $\sigma^2 + \|\theta\|^2/k$. In this case, the KL term is no less than 
\begin{align*}
 k \log\Big(1 +\frac{\|\theta_0\|^2}{k\sigma^2}  \Big).
\end{align*}
Thus, the second term of RHS is 
\begin{align*}
    \sqrt{\frac{\mathsf{KL}(Q\|P) + \log \frac{1}{\delta}}{2N}}\times L \geq \sqrt{\frac{k\log\big(1 + \frac{\|\theta\|^2}{k\sigma^2} \big)}{4N}}\times L \geq L
\end{align*}
when $\|\theta\|^2 > \sigma^2 \big\{\exp(4N/k)-1\big\}$. Hence, for any $\|\theta\|_2 > \sigma^2 \big\{\exp(4N/k)-1 \big\}$, we have the RHS is greater than the LHS, the inequality is trivial. In this work, we only consider the case: 
\begin{align}\label{cond:theta}\|\theta\|^2 < \sigma^2 \big(\exp\{4N/k\} -1\big).
\end{align}
Distribution $P$ is Gaussian centered around  $\mathbf{0}$ with variance $\sigma_P^2 = \sigma^2 + \|\theta\|^2/k$, which is unknown at the time we set up the inequality, since $\theta$ is unknown. Meanwhile we have to specify $P$ in advance, since $P$ is the prior distribution. To deal with this problem, we could choose a family of $P$ such that its means cover the space of $\theta$ satisfying inequality \eqref{cond:theta}. We set
\begin{align*}
c&= \sigma^2\big(1 + \exp\{4N/k\}\big)\\
P_j &= \mathcal{N}\big(0,c\exp\frac{1-j}{k}\mathbb{I}_k\big)\\
    \mathfrak{P}&:= \big\{P_j: j = 1,2,\ldots \big\}
\end{align*}
Then the following inequality holds for a particular distribution $P_j$ with probability $1-\delta_j$ with $\delta_j = \frac{6\delta}{\pi^2 j^2}$
\begin{align*}
    \mathbb{E}_{\theta^{\prime}\sim \mathcal{N}(\theta,\sigma^2)}\Lc_{\Dc}\big(f_{\theta^{\prime}} \big)&\leq \mathbb{E}_{\theta^{\prime}\sim \mathcal{N}(\theta,\sigma^2)} \Lc_{\Sc}\big(f_{\theta^{\prime}}\big) + \frac{1}{\beta}\left[  \mathsf{KL}(Q\|P_j) + \log\frac{1}{\delta_j} + \Psi(\beta,N) \right].
\end{align*}
Use the well-known equation: $\sum_{j=1}^{\infty} \frac{1}{j^2} = \frac{\pi^2}{6}$, then with probability $1-\delta$, the above inequality holds with every $j$. We pick
\begin{align*}
j^*:= \left\lfloor 1- k \log\frac{\sigma^2 + \|\theta\|^2/k}{c} \right\rfloor = \left\lfloor 1- k \log\frac{\sigma^2 + \|\theta\|^2/k}{\sigma^2(1+ \exp\{4N/k\})} \right\rfloor.
\end{align*}
Therefore,
\begin{align*}
&1 - j^* = \left\lceil k \log \frac{\sigma^2 + \|\theta\|^2/k}{c}\right\rceil \\
\Rightarrow \quad &\log \frac{\sigma^2 + \|\theta\|^2/k}{c}\leq \frac{1-j^*}{k} \leq \log\frac{\sigma^2 + \|\theta_0\|^2/k}{c} + \frac{1}{k}\\
\Rightarrow \quad & \sigma^2 + \|\theta\|^2/k \leq c\exp\left\{\frac{1-j^*}{k} \right\} \leq \exp(1/k) \big[\sigma^2 + \|\theta\|^2/k \big]\\
\Rightarrow \quad & \sigma^2 + \|\theta\|^2/k \leq \sigma_{P_{j^*}}^2\leq \exp(1/k) \big[\sigma^2 + \|\theta\|^2/k \big].
\end{align*}
Thus the KL term could be bounded as follow
\begin{align*}
    \mathsf{KL}(Q\|P_{j^*}) &= \frac{1}{2}\left[\frac{k\sigma^2 + \|\theta\|^2}{\sigma_{P_{j^*}}^2}- k + k \log \frac{\sigma_{P_{j^*}}^2}{\sigma^2} \right]\\
    &\leq \frac{1}{2}\left[\frac{k(\sigma^2 + \|\theta\|^2/k)}{\sigma^2 + \|\theta\|^2/k} - k + k \log \frac{\exp(1/k)\big(\sigma^2 + \|\theta\|^2/k\big)}{\sigma^2} \right] \\
    &= \frac{1}{2}\Big[k \log \frac{\exp(1/k)\big(\sigma^2 + \|\theta\|^2/k \big)}{\sigma^2} \Big] \\
    &= \frac{1}{2}\Big[1 + k\log\Big(1 + \frac{\|\theta_0\|^2}{k\sigma^2} \Big) \Big]
\end{align*}
For the term $\log \frac{1}{\delta_{j^*}}$, with recall that $c = \sigma^2\big(1+\exp(4N/k) \big)$ and

$j^* = \left\lfloor 1- k \log\frac{\sigma^2 + \|\theta\|^2/k}{\sigma^2(1+ \exp\{4N/k\})} \right\rfloor$, we have
\begin{align*}
    \log\frac{1}{\delta_{j^*}} &= \log \frac{(j^*)^2\pi^2}{6\delta}  = \log\frac{1}{\delta}  + \log\Big(\frac{\pi^2}{6}\Big) + 2\log(j^*) \\
    &\leq \log\frac{1}{\delta} + \log\frac{\pi^2}{6} + 2\log \Big( 1+k\log\frac{\sigma^2\big(1+ \exp(4N/k)\big)}{\sigma^2 + \|\theta\|^2/k}\Big)  \\
    &\leq \log\frac{1}{\delta} + \log\frac{\pi^2}{6} + 2\log\Big(1+ k\log\big(1+\exp(4N/k)\big)\Big) \\
    &\leq \log\frac{1}{\delta} + \log\frac{\pi^2}{6} + 2\log\Big(1+ k\big(1+\frac{4N}{k} \big) \Big) \\
    &\leq \log\frac{1}{\delta} + \log\frac{\pi^2}{6} + \log(1+k + 4N).
\end{align*}
Hence, the inequality 
\begin{align*}
    \Lc_{\Dc}\Big(\theta^{\prime},\mathcal{N}(\theta,\sigma^2\mathbb{I}_k)\Big)&\leq \Lc_{\Sc}\Big(\theta^{\prime},\mathcal{N}(\theta,\sigma^2\mathbb{I}_k) \Big) + \sqrt{\frac{\mathsf{KL}(Q\|P_{j^*}) + \log \frac{1}{\delta_{j^*}}}{2N}}\times L \\
    &\leq \Lc_{\Sc}\Big(\theta^{\prime},\mathcal{N}(\theta,\sigma^2\mathbb{I}_k) \Big) \\ 
    & + \frac{L}{2\sqrt{N}}\sqrt{1 + k\log\Big(1+ \frac{\|\theta\|^2}{k\sigma^2}\Big) + 2 \log \frac{\pi^2}{6\delta} + 4 \log(N+k)}\\
    &\leq  \Lc_{\Sc}\Big(\theta^{\prime},\mathcal{N}(\theta,\sigma^2\mathbb{I}_k)\Big) \\ 
    & + \frac{L}{2\sqrt{N}} \sqrt{k\log\big(1+ \frac{\|\theta\|^2}{k\sigma^2}\big)+ O(1) + 2\log\frac{1}{\delta} + 4\log(N+k)}.
\end{align*}
Since $\|\theta^{\prime}-\theta\|^2$ is $k$ chi-square distribution, for any positive $t$,
we have
\begin{align*}
    \mathbb{P}\big(\|\theta^{\prime}-\theta\|^2 - k \sigma^2 \geq 2\sigma^2 \sqrt{kt} + 2t\sigma^2\big) \big) \leq \exp(-t).
\end{align*}
By choosing $t = \frac{1}{2}\log(N)$, with probability $1-N^{-1/2}$, we have
\begin{align*}
    \|\theta^{\prime}-\theta\|^2 \leq \sigma^2 \log(N) + k\sigma^2 + \sigma^2\sqrt{2 k\log(N)} \leq k\sigma^2 \Big(1 + \sqrt{\frac{\log(N)}{k}} \Big)^2.
\end{align*}
 By setting $\sigma = \rho\times \big(\sqrt{k} + \sqrt{\log(N)}\big)^{-1}$, we have $\|\theta^{\prime}-\theta\|^2 \leq \rho^2$. Hence, we get
 \begin{align*}
     \Lc_{\Sc}\Big(\theta^{\prime},\mathcal{N}(\theta,\sigma^2\mathbb{I}_k)\Big) &= \mathbb{E}_{\theta\sim\mathcal{N}(\theta,\sigma^2\mathbb{I}_k)}\mathbb{E}_{\Sc}\big[f_{\theta^{\prime}} \big] = \int_{\|\theta^{\prime}-\theta\|\leq \rho} \mathbb{E}_{\Sc}\big[f_{\theta^{\prime}}\big]d\mathcal{N}(\theta,\sigma^2\mathbb{I}) \\ 
     & + \int_{\|\theta^{\prime}-\theta\|> \rho} \mathbb{E}_{\Sc}\big[f_{\theta^{\prime}} \big] d\mathcal{N}(\theta,\sigma^2\mathbb{I})\\
     &\leq \Big(1-\frac{1}{\sqrt{N}} \Big)\max_{\|\theta^{\prime} - \theta\|\leq \rho} \Lc_{\Sc}(\theta^{\prime}) + \frac{1}{\sqrt{N}}L \\
     &\leq \max_{\|\theta^{\prime}-\theta\|_2 \leq \rho} \Lc_{\Sc}(\theta^{\prime}) + \frac{2L}{\sqrt{N}}.
 \end{align*}
 It follows that
\begin{align*}
    \Lc_{\Dc}(\theta) \leq \max_{\|\theta^{\prime}-\theta\|\leq \rho} \Lc_{\Sc}(\theta^{\prime}) & +\frac{4L}{\sqrt{N}}\Bigg[\sqrt{k\log\Big(1 + \frac{\|\theta\|^2}{\rho^2} \big(1+\sqrt{\log(N)/k}\big)^2 \Big)} \\ 
    & + 2\sqrt{\log\big(\frac{N+k}{\delta}\big)} + O(1) \Bigg].
\end{align*}
Replace $\theta = \theta_i$, with probability $1-\delta/(m+1)$ we have
\begin{align*}
    \Lc_{\Dc}(\theta_i) & \leq \Lc_{\Dc}\Big(\theta_i^{\prime},\Nc\big(\theta_i,\sigma^2\mathbb{I}\big) \Big) \leq \max_{\|\theta_i^{\prime} - \theta_i\|<\rho}\Lc_{\Sc}(\theta_i^{\prime}) \\ 
    & + \frac{4L}{\sqrt{N}} \Bigg[\sqrt{k\log\Big(1+\frac{\|\theta_i\|^2}{\rho^2}\big(1+\sqrt{\log(N)/k}\big)^2\Big) } \\
    & + 2\sqrt{\log\frac{(m+1)(N+k)}{\delta}}+ O(1) \Bigg].
\end{align*}
For the loss on the ensemble classifier, we use the assumption
\begin{align*}
    \mathbb{E}_{(x,y)\in \mathcal{D}}\big[\ell( f_{\theta}^{\mathrm{ens}}(x),y) \big] &\leq  \mathbb{E}_{\theta \sim P}\mathbb{E}_{(x,y)\in\mathcal{D}} \big[\ell(f_{\theta}^{\mathrm{ens}}(x),y)\big].
\end{align*}
Repeating the same step of proof for $\theta$, with probability at least $1- \delta/(m+1)$, we obtain
\begin{align*}
    \Lc_{\Dc}(\theta) &\leq \Lc_{\Dc}\big(\theta^{\prime},\Nc(\theta,\sigma^2\mathbb{I}_{mk})\big) \\
    &\leq \max_{\|\theta^{\prime} - \theta\|\leq \sqrt{m}\rho}{\Lc}_{\Sc}(\theta^{\prime}) \\ 
    & + \frac{4L}{\sqrt{N}} \Bigg[\sqrt{mk\log\Big(1+ \frac{\sum_{i=1}^m\|\theta_i\|^2}{m\rho^2}  \big(1+\sqrt{\log(N)/mk} \big)^2\Big)} + \\
    & \qquad \qquad 2 \sqrt{ \log\frac{(m+1)(N+mk)}{\delta}} + O(1)\Bigg].
\end{align*}
By the convexity property of $\ell$, we have
\begin{align*}
    \Lc_{\Dc}(\theta) &\leq \E_{(x,y)\sim \Dc}\Big[\ell\Big( \frac{1}{m}\sum_{i=1}^m f_{\theta_i}(x)^{(i)},y\Big) \Big] \leq \E_{(x,y)\sim \Dc} \Big[\frac{1}{m}\sum_{i=1}^m \ell\big(f_{\theta_i}^{(i)}(x),y \big) \Big] \\
    &\leq \frac{1}{m}\sum_{i=1}^m \Lc_{\Dc}(\theta_i).
\end{align*}
Finally, we obtain
\begin{align*}
    \Lc_{\Dc}(\theta) &\leq \gamma \max_{\theta^{\prime}: \|\theta^{\prime} -\theta\|\leq \sqrt{m} \rho} \Lc_{\Sc}(\theta^{\prime}) + \frac{1-\gamma}{m} \Big[\sum_{i=1}^m\max_{\theta^{\prime}_i:\|\theta^{\prime}_i - \theta_i\|\leq \rho} \Lc_{\Sc}(\theta^{\prime}_i)\Big] +  
   \frac{CL}{\sqrt{N}} \times\\ &\qquad \Bigg[m\sqrt{\log\frac{m(N+k)}{\delta} } + \sum_{i=1}^m \sqrt{k\log\Big(1+\frac{\|\theta_i\|^2}{\rho^2}(1+\sqrt{\log(N)}/k)^2\Big)} \\
    & + \sqrt{km \log\Big(1+ \frac{\sum_{i=1}^m \|\theta_i\|^2}{m\rho^2}\big(1+ \sqrt{\log(N)/(mk)}\big)^2 \Big)} + O(1)\Bigg]
\end{align*}
where $C$ is an universal constant.
\end{proof}

\section{Training Algorithm}

We present the pseudo code for our proposed method DASH as in Algorithm \ref{alg:dash_algo}. It is worth noting that we utilize the cross entropy loss with label smoothing with $\alpha=0.1$ as the loss function $l$. Compared to the standard SAM ~\cite{foret2021sharpnessaware}, our method has one modification in the first optimization phase when we consider the diverse-aware loss $\mathcal{L}_{B}^{div} (\theta_i, \theta_{\neq i}) $ in addition to the predictive loss $\widetilde{\mathcal{L}_{B}}(\theta_i)$ in order to find the perturbed weight $\theta_i^a$. 
However, this process requires to calculate the gradients $\nabla_{\theta_i} \mathcal{L}_{B}(\theta_i)$ and $\nabla_{\theta_i} \mathcal{L}_{B}^{div}(\theta_i,\theta_{\neq i})$ of the two losses with respect to the same $\theta_i$  separately which consumes one more back-propagation step compared to SAM. Therefore, in practice, we alternatively consider to maximize the combined loss $\mathcal{L}_{B}^{c} (\theta_i) = \widetilde{\mathcal{L}_{B}} (\theta_i) + \gamma_c \; \mathcal{L}_{B}^{div} (\theta_i, \theta_{\neq i})$ to find the perturbed weight $\theta_i^a$. The trade-off parameter $\gamma_c$ now replaces the perturbed radius $\rho_2$ and can be found adaptively by adjusting the strength of two gradients in the first iteration of each epoch.  By using this approach, we can use the same number of back-propagation step as SAM in the first optimization phase.

\begin{algorithm}
\caption{DASH Algorithm}
\label{alg:dash_algo}
\textbf{Input:} Training set $\mathcal{S} \triangleq \{(x_n, y_n)\}^{N}_{n=1}$; Loss function $l: \mathbb{R}^M \times \mathcal{Y}$; Batch size $b$; Learning rate $\eta$; Trade-off parameter $\gamma$; Perturbed radiuses $\rho_1, \rho_2$; Ensemble size $m$. 

\textbf{Output:} Ensemble trained with DASH $\theta^{t}$\\

Initialize weights for $m$ base learners $\theta^0 := [\theta_i]_{i=1}^{m}, t = 0$; 

\textbf{While} \textit{not converged} \textit{do}
    
    \quad Sample batch $B = \{ (x_1, y_1), ...(x_b, y_b)\}$;
    
    \quad \textit{For} $i \leftarrow 1$ to $m$ \textit{do}
    
    \quad \quad Compute gradient $\nabla_{\theta_{i}}\widetilde{\mathcal{L}_{B}}\left(\theta_{i}\right)$ of the batch's training loss per Eq. (3);

    \quad \quad Compute $i$-th divergence loss $\mathcal{L}_{B}^{div}\left(\theta_{i},\theta_{\neq i}\right)$ per Eq. (5);
        
    \quad \quad     Compute the perturbed weight per Eq. (6); 
    $$\theta_{i}^{a} = \theta_{i}+\rho_{1}\frac{\nabla_{\theta_{i}}\widetilde{\mathcal{L}_{B}}\left(\theta_{i}\right)}{\Vert\nabla_{\theta_{i}}\widetilde{\mathcal{L}_{B}}\left(\theta_{i}\right)\Vert}+\rho_{2}\frac{\nabla_{\theta_{i}}\mathcal{L}_{B}^{div}\left(\theta_{i},\theta_{\neq i}\right)}{\Vert\nabla_{\theta_{i}}\mathcal{L}_{B}^{div}\left(\theta_{i},\theta_{\neq i}\right)\Vert}$$
        
    \quad \quad     Update weights: $\theta_{i} = \theta_{i} - \eta \nabla_{\theta_{i}}\mathcal{L}_{B}\left(\theta_{i}^{a}\right)$;\\
    
    \quad $\theta^{t+1} \gets [\theta_{i}]^{m}_{i}$;

    \quad $t = t + 1$

\end{algorithm}

\section{Additional Experiments}

\subsection{Evaluation of Uncertainty Estimation}

We would like to provide the complete experimental results with all \textbf{six} Uncertainty Estimation metrics on the Tiny-ImageNet dataset (Table \ref{tab:eval-tiny-r18x3}), the CIFAR10 dataset (Tables \ref{tab:eval-c10-r10x5}, \ref{tab:eval-c10-r18x3}, \ref{tab:eval-c10-rme}) and the CIFAR100 dataset (Tables \ref{tab:eval-c100-r10x5}, \ref{tab:eval-c100-r18x3}, \ref{tab:eval-c100-rme}).

On evaluation of the predictive performance, as reported in Section 4.3 in the main paper, our proposed method DASH consistently and significantly outperforms all baselines across all datasets and architectures. Unlike Fast Geometric, Snapshot, or EDST methods, which are limited to homogeneous ensemble setting, our DASH is a general method capable on either homogeneous or heterogeneous ensemble.

On evaluation of the uncertainty estimation capability, in addition to the result on the Tiny-Imagenet dataset that has been reported in Section 4.4 in the main paper, Tables \ref{tab:eval-c10-r10x5}, \ref{tab:eval-c10-r18x3}, \ref{tab:eval-c10-rme} show the results on the CIFAR10 dataset, with R10x5, R18x3 and RME architectures, respectively, while Tables \ref{tab:eval-c100-r10x5}, \ref{tab:eval-c100-r18x3}, \ref{tab:eval-c100-rme} show the results of the same architectures on the CIFAR100 dataset. 
As similar as the observation on the Tiny-Imagenet dataset, it can be seen from the results on the CIFAR10 and CIFAR100 datasets that our proposed method DASH achieves the best performance on the five UE metrics, except for the ECE metric. In comparison to the Deep Ensemble, our method achieves much better performance on the heterogeneous setting (i.e., RME architecture) as seen from Table \ref{tab:eval-c10-rme} or Table \ref{tab:eval-c100-rme}. Overall, the experimental results demonstrate the superiority of our proposed DASH method over other baseline methods for both predictive performance and UE capabilities across all datasets and architectures.

\begin{center}
\par\end{center}

\begin{center}
\begin{table}[!ht]
\begin{centering}
\caption{Evaluation on the Tiny-ImageNet dataset with R18x3 architecture.\label{tab:eval-tiny-r18x3}}
\par\end{centering}
\begin{centering}
\begin{tabular}{lccccccc}
\toprule
 & Accuracy  $\uparrow$ & NLL $\downarrow$ & Brier $\downarrow$ & ECE $\downarrow$ & Cal-Brier $\downarrow$ & Cal-AAC $\downarrow$ & Cal-NLL $\downarrow$\tabularnewline
\midrule 
Deep Ensemble & 65.9 & 1.400 & 0.452 & \textbf{0.110} & 0.453 & 0.210 & 1.413\tabularnewline
Fast Geometric & 61.8 & 1.548 & 0.501 & 0.116 & 0.499 & 0.239 & 1.544\tabularnewline
Snapshot & 62.2 & 1.643 & 0.505 & 0.118 & 0.501 & 0.237 & 1.599\tabularnewline
EDST & 62.3 & 1.581 & 0.496 & 0.115 & 0.495 & 0.235 & 1.548\tabularnewline
DST & 61.9 & 1.525 & 0.499 & \textbf{0.110} & 0.500 & 0.239 & 1.536\tabularnewline
SGD & 62.3 & 1.999 & 0.601 & 0.283 & 0.518 & 0.272 & 1.737\tabularnewline
SAM & 66.1 & 1.791 & 0.563 & 0.297 & 0.469 & 0.242 & 1.484\tabularnewline
DASH (Ours) & \textbf{69.9} & \textbf{1.379} & \textbf{0.447} & 0.184 & \textbf{0.407} & \textbf{0.204} & \textbf{1.213}\tabularnewline
\bottomrule 
\end{tabular}
\par\end{centering}
\centering{}
\end{table}
\par\end{center}

\begin{center}
\begin{table}[!ht]
\begin{centering}
\caption{Evaluation on the CIFAR10 dataset with R10x5 architecture.\label{tab:eval-c10-r10x5}}
\par\end{centering}
\begin{centering}
\begin{tabular}{lccccccc}
\toprule
 & Accuracy   $\uparrow$ & NLL $\downarrow$ & Brier $\downarrow$ & ECE $\downarrow$ & Cal-Brier $\downarrow$ & Cal-AAC $\downarrow$ & Cal-NLL $\downarrow$\tabularnewline
\midrule 
Deep Ensemble & 92.7 & 0.226 & 0.107 & \textbf{0.053} & 0.091 & 0.108 & 0.272\tabularnewline
Fast Geometric & 92.5 & 0.555 & 0.261 & 0.113 & 0.251 & 0.144 & 0.531\tabularnewline
Snapshot & 93.6 & 0.202 & 0.095 & 0.048 & 0.083 & \textbf{0.107} & 0.249\tabularnewline
EDST & 92.0 & 0.245 & 0.118 & 0.057 & 0.122 & 0.112 & 0.301\tabularnewline
DST & 93.2 & 0.211 & 0.099 & 0.049 & 0.102 & 0.108 & 0.261\tabularnewline
SGD & 95.1 & 0.277 & 0.096 & 0.143 & 0.078 & 0.108 & 0.264\tabularnewline
SAM & 95.4 & 0.257 & 0.087 & 0.136 & 0.073 & \textbf{0.107} & 0.268\tabularnewline
DASH (Ours) & \textbf{95.7} & \textbf{0.244} & \textbf{0.084} & 0.134 & \textbf{0.067} & \textbf{0.107} & \textbf{0.248}\tabularnewline
\hline 
\end{tabular}
\par\end{centering}
\centering{}
\end{table}
\par\end{center}

\begin{center}
\begin{table}[!ht]
\begin{centering}
\caption{Evaluation on the CIFAR10 dataset with R18x3 architecture.\label{tab:eval-c10-r18x3}}
\par\end{centering}
\begin{centering}
\begin{tabular}{lccccccc}
\toprule
 & Accuracy  $\uparrow$ & NLL $\downarrow$ & Brier $\downarrow$ & ECE $\downarrow$ & Cal-Brier $\downarrow$ & Cal-AAC $\downarrow$ & Cal-NLL $\downarrow$\tabularnewline
\midrule 
Deep Ensemble & 93.7 & 0.197 & 0.091 & 0.047 & 0.079 & \textbf{0.107} & 0.273\tabularnewline
Fast Geometric & 93.3 & 0.257 & 0.108 & 0.055 & 0.087 & 0.108 & 0.261\tabularnewline
Snapshot & 94.8 & 0.201 & 0.082 & 0.043 & 0.071 & 0.108 & 0.270\tabularnewline
EDST & 92.8 & 0.231 & 0.110 & 0.054 & 0.113 & 0.110 & 0.281\tabularnewline
DST & 94.7 & 0.172 & 0.080 & \textbf{0.042} & 0.083 & \textbf{0.107} & 0.253\tabularnewline
SGD & 95.2 & 0.249 & 0.083 & 0.120 & 0.076 & 0.108 & 0.282\tabularnewline
SAM & 95.8 & 0.229 & 0.074 & 0.120 & 0.067 & \textbf{0.107} & 0.261\tabularnewline
DASH (Ours) & \textbf{96.7} & \textbf{0.215} & \textbf{0.065} & 0.124 & \textbf{0.056} & \textbf{0.107} & \textbf{0.250}\tabularnewline
\bottomrule 
\end{tabular}
\par\end{centering}
\centering{}
\end{table}
\par\end{center}

\begin{center}
\begin{table}[!ht]
\begin{centering}
\caption{Evaluation on the CIFAR10 dataset with RME architecture.\label{tab:eval-c10-rme}}
\begin{tabular}{lccccccc}
\toprule
 & Accuracy  $\uparrow$ & NLL $\downarrow$ & Brier $\downarrow$ & ECE $\downarrow$ & Cal-Brier $\downarrow$ & Cal-AAC $\downarrow$ & Cal-NLL $\downarrow$\tabularnewline
\midrule 
Deep Ensemble & 89.0 & 0.905 & 0.391 & 0.431 & 0.153 & 0.126 & 0.395\tabularnewline
DST & 93.4 & 0.209 & 0.101 & \textbf{0.058} & 0.102 & 0.109 & 0.282\tabularnewline
SGD & 92.6 & 0.328 & 0.128 & 0.136 & 0.113 & 0.112 & 0.317\tabularnewline
SAM & 93.8 & 0.310 & 0.112 & 0.145 & 0.094 & 0.110 & 0.280\tabularnewline
DASH (Ours) & \textbf{95.2} & \textbf{0.276} & \textbf{0.095} & 0.151 & \textbf{0.075} & \textbf{0.106} & \textbf{0.236}\tabularnewline
\bottomrule 
\end{tabular}
\par\end{centering}
\centering{}
\end{table}
\par\end{center}

\begin{center}
\begin{table}[!ht]
\begin{centering}
\caption{Evaluation on the CIFAR100 dataset with R10x5 architecture.\label{tab:eval-c100-r10x5}}
\begin{tabular}{lccccccc}
\toprule
 & Accuracy  $\uparrow$ & NLL $\downarrow$ & Brier $\downarrow$ & ECE $\downarrow$ & Cal-Brier $\downarrow$ & Cal-AAC $\downarrow$ & Cal-NLL $\downarrow$\tabularnewline
\midrule 
Deep Ensemble & 73.7 & 0.973 & 0.365 & \textbf{0.101} & 0.329 & 0.162 & 0.870\tabularnewline
Fast Geometric & 63.2 & 1.926 & 0.658 & 0.213 & 0.606 & 0.324 & 1.723\tabularnewline
Snapshot & 72.8 & 1.072 & 0.382 & 0.112 & 0.338 & 0.165 & 0.929\tabularnewline
EDST & 68.4 & 1.142 & 0.427 & 0.112 & 0.427 & 0.207 & 1.151\tabularnewline
DST & 70.8 & 1.064 & 0.393 & 0.103 & 0.396 & 0.189 & 1.076\tabularnewline
SGD & 75.9 & 1.502 & 0.522 & 0.400 & 0.346 & 0.174 & 1.001\tabularnewline
SAM & 77.7 & 1.302 & 0.460 & 0.357 & 0.321 & 0.164 & 0.892\tabularnewline
DASH (Ours) & \textbf{80.8} & \textbf{0.864} & \textbf{0.316} & 0.180 & \textbf{0.271} & \textbf{0.144} & \textbf{0.684}\tabularnewline
\bottomrule 
\end{tabular}
\par\end{centering}
\centering{}
\end{table}
\par\end{center}

\begin{center}
\begin{table}[!ht]
\begin{centering}
\caption{Evaluation on the CIFAR100 dataset with R18x3 architecture.\label{tab:eval-c100-r18x3}}
\begin{tabular}{lccccccc}
\toprule
 & Accuracy  $\uparrow$ & NLL $\downarrow$ & Brier $\downarrow$ & ECE $\downarrow$ & Cal-Brier $\downarrow$ & Cal-AAC $\downarrow$ & Cal-NLL$\downarrow$\tabularnewline
\midrule 
Deep Ensemble & 75.4 & 0.927 & 0.342 & \textbf{0.095} & 0.308 & 0.155 & 0.822\tabularnewline
Fast Geometric & 72.3 & 1.12 & 0.394 & 0.124 & 0.344 & 0.169 & 0.950\tabularnewline
Snapshot & 75.7 & 1.011 & 0.347 & 0.111 & 0.311 & 0.153 & 0.903\tabularnewline
EDST & 69.6 & 1.125 & 0.412 & 0.106 & 0.412 & 0.197 & 1.123\tabularnewline
DST & 70.4 & 1.228 & 0.419 & 0.140 & 0.405 & 0.194 & 1.153\tabularnewline
SGD & 78.9 & 1.225 & 0.389 & 0.285 & 0.304 & 0.156 & 0.919\tabularnewline
SAM & 80.1 & 1.080 & 0.356 & 0.261 & 0.285 & 0.151 & 0.808\tabularnewline
DASH (Ours) & \textbf{82.2} & \textbf{0.892} & \textbf{0.300} & 0.196 & \textbf{0.255} & \textbf{0.138} & \textbf{0.679}\tabularnewline
\hline 
\end{tabular}
\par\end{centering}
\centering{}
\end{table}
\par\end{center}

\begin{center}
\begin{table}[!ht]
\begin{centering}
\caption{Evaluation on the CIFAR100 dataset with RME architecture.\label{tab:eval-c100-rme}}
\begin{tabular}{lccccccc}
\toprule
 & Accuracy $\uparrow$ & NLL$\downarrow$ & Brier $\downarrow$ & ECE $\downarrow$ & Cal-Brier $\downarrow$ & Cal-AAC $\downarrow$ & Cal-NLL $\downarrow$\tabularnewline
\midrule 
Deep Ensemble & 62.7 & 2.137 & 0.699 & 0.401 & 0.433 & 0.209 & 1.267\tabularnewline
DST & 71.7 & 1.056 & 0.393 & \textbf{0.111} & 0.393 & 0.187 & 1.066\tabularnewline
SGD & 72.6 & 1.559 & 0.531 & 0.350 & 0.403 & 0.201 & 1.192\tabularnewline
SAM & 76.4 & 1.439 & 0.501 & 0.377 & 0.347 & 0.177 & 1.005\tabularnewline
DASH (Ours) & \textbf{78.7} & \textbf{0.969} & \textbf{0.342} & 0.202 & \textbf{0.298} & \textbf{0.151} & \textbf{0.764}\tabularnewline
\bottomrule 
\end{tabular}
\par\end{centering}
\centering{}
\end{table}
\par\end{center}

\subsection{Effect of Sharpness-aware Minimization}

Since proposed in ~\cite{foret2021sharpnessaware}, there are several sharpness-aware minimization methods have been developed to address various limitations of the pioneer method. Notably, ~\cite{kwon2021asam} proposed an adaptive method to reduce the sensitivity to parameter re-scaling issue, thus reducing the gap between sharpness and generalization of a model. In this section, we would like to examine the impact of different sharpness-aware methods to the final performance when integrating into our method. More specifically, we consider two sharpness-aware methods which are Standard (Non-Adaptive) SAM ~\cite{foret2021sharpnessaware} and Adaptive SAM ~\cite{kwon2021asam}, corresponding to our two variants which are Standard DASH and Adaptive DASH. We conduct experiment on the CIFAR10 and CIFAR100 datasets with two ensemble settings, i.e., R18x3 and R10x5 architectures and report results in Table \ref{tab:abl-adaptive-sam}. We choose $\rho=0.05$ for the standard version and $\rho=2.0$ for the adaptive version as suggested in the project \footnote{\url{https://github.com/davda54/sam}}. 
The results show that integrating sharpness-aware methods into our approach yields a significant improvement, regardless of the version. For example, Adaptive-DASH outperforms Adaptive-SAM across all settings, in both generalization and uncertainty estimation capability. Notably, the improvements are 1.3\% and 1.76\% in CIFAR100 dataset prediction tasks with R18x3 and R10x5 architectures, respectively. Similarly, Standard-DASH achieves a significant improvement over Standard-SAM in all settings, with the highest improvement being 2.55\% ensemble accuracy with the R10x5 architecture on the CIFAR100 dataset. Interestingly, our Standard-DASH version even outperforms the Adaptive-SAM to achieve the second-best performance, just after our Adaptive-DASH version. This result emphasizes the effectiveness and generality of our method in various settings. Based on these results, we use the Adaptive-DASH as the default setting.

\begin{table}
\caption{Analysis of the effect of the sharpness aware methods on the CIFAR10 (C10) and CIFAR100 (C100) datasets. "A" denotes the adaptive sharpness-aware minimization, which is scale-invariant as proposed in. "S" denotes the standard (non adaptive) version.\label{tab:abl-adaptive-sam}}
\begin{centering}
\resizebox{0.75\columnwidth}{!}{%
\begin{tabular}{llccccc}
 \toprule
 &  & \multicolumn{2}{c}{R18x3} &  & \multicolumn{2}{c}{R10x5}\tabularnewline
\cline{3-4} \cline{4-4} \cline{6-7} \cline{7-7} 
\noalign{\vspace{0.5ex}} 
 &  & Accuracy $\uparrow$ & Cal-Brier $\downarrow$ &  & Accuracy $\uparrow$ & Cal-Brier $\downarrow$\tabularnewline
\midrule
\multirow{5}{*}{C10} & SGD & 95.20 & 0.076 &  & 95.06 & 0.078\tabularnewline
 & SAM & 95.80 & 0.067 &  & 95.43 & 0.073\tabularnewline
 & S-DASH & \underline{96.38} & \underline{0.060} &  & \textbf{95.80} & \textbf{0.066}\tabularnewline
 & A-SAM & 96.02 & 0.064 &  & 95.65 & 0.071\tabularnewline
 & A-DASH & \textbf{96.66} & \textbf{0.056} &  & \underline{95.72} & \underline{0.067}\tabularnewline
\midrule
\multirow{5}{*}{C100} & SGD & 78.93 & 0.304 &  & 75.85 & 0.346\tabularnewline
 & SAM & 80.07 & 0.285 &  & 77.71 & 0.321\tabularnewline
 & S-DASH & \underline{81.73} & \underline{0.262} &  & \underline{80.26} & \underline{0.279}\tabularnewline
 & A-SAM & 80.89 & 0.275 &  & 79.08 & 0.301\tabularnewline
 & A-DASH & \textbf{82.19} & \textbf{0.255} &  & \textbf{80.84} & \textbf{0.267}\tabularnewline
\hline 
\end{tabular}}
\par\end{centering}
\centering{}
\end{table}

\subsection{Analysis of the ensemble size}

We here examine the impact of ensemble size, that is, the number of base classifiers, on the final performance. 
We performed an experiment on the CIFAR100 dataset by varying the ensemble size from two to seven base classifiers, in which each base classifier is a ResNet10 model. 
The results of this experiment are presented in Table \ref{tab:abl-num_models}.
Ensemble learning theory suggests that the generalization capacity of an ensemble improves with the number of base classifiers, assuming the base classifiers exhibit diversity. 
Table \ref{tab:abl-num_models} demonstrates that the ensemble accuracy increases linearly with the number of base classifiers, with a $1.7\%$ improvement in ensemble accuracy when increasing the number of base classifiers from $2$ to $6$. 
Furthermore, our method's uncertainty estimation capability benefits from a larger ensemble size, as evidenced by the improvements in all three UE metrics. Interestingly, we also observed that the performance of the base classifiers, as measured by the average accuracy metric in Table \ref{tab:abl-num_models}, also improves when working in collaboration with a larger number of base classifiers, with an accuracy improvement of $0.4\%$. However, we noted that the benefits of a larger ensemble appear to reach a saturation point when the number of base classifiers exceeds $6$.   

\begin{table}[h!]
\caption{Ensemble performance on CIFAR100 with different number of base classifiers. }
\label{tab:abl-num_models}
\centering
\begin{tabular}{lccccc}
\toprule
 & Accuracy $\uparrow$ & Avg. Accuracy   $\uparrow$ & Cal-Brier $\downarrow$ & Cal-AAC $\downarrow$ & Cal-NLL $\downarrow$\tabularnewline
\midrule
R10x2 & 79.19 & 77.60 & 0.289 & 0.792 & 0.751\tabularnewline
R10x3 & 79.86 & 77.49 & 0.280 & 0.147 & 0.715\tabularnewline
R10x4 & 80.71 & 77.77 & 0.272 & 0.143 & 0.691\tabularnewline
R10x5 & 80.84 & 77.94 & 0.267 & 0.142 & 0.676\tabularnewline
R10x6 & 80.89 & 77.98 & 0.268 & 0.142 & 0.677\tabularnewline
R10x7 & 80.89 & 77.88 & 0.268 & 0.142 & 0.673\tabularnewline
\bottomrule 
\end{tabular}
\end{table}

\bibliographystyle{splncs04}
\bibliography{dash}